\documentclass[10pt,twocolumn,letterpaper]{article}

\usepackage[pagenumbers]{cvpr} % To force page numbers, e.g. for an arXiv version

\definecolor{lightblue}{RGB}{217, 230, 224}
\definecolor{forestgreen}{RGB}{74, 124, 89}
\usepackage{color}
\usepackage{amsmath}
\usepackage{graphicx}
\usepackage{booktabs}
\usepackage{url}
\usepackage{wrapfig}
\usepackage{algorithm}
\usepackage{algpseudocode}

\usepackage{indentfirst}
\usepackage{footnote}
\usepackage{multirow}
\usepackage{pifont}

\usepackage{colortbl}
\usepackage{soul}  % background text color

\usepackage{cuted}
\usepackage{caption}

\usepackage{amsmath}
\definecolor{commentcolor}{RGB}{110,154,155}  % define a teal-gray

\usepackage{amssymb}
\newcommand{\cmark}{\checkmark}

\definecolor{cvprblue}{rgb}{0.21,0.49,0.74}
\usepackage[pagebackref,breaklinks,colorlinks,allcolors=cvprblue]{hyperref}

\title{EarlyTom\raisebox{-0.35em}{\includegraphics[width=0.07\textwidth]{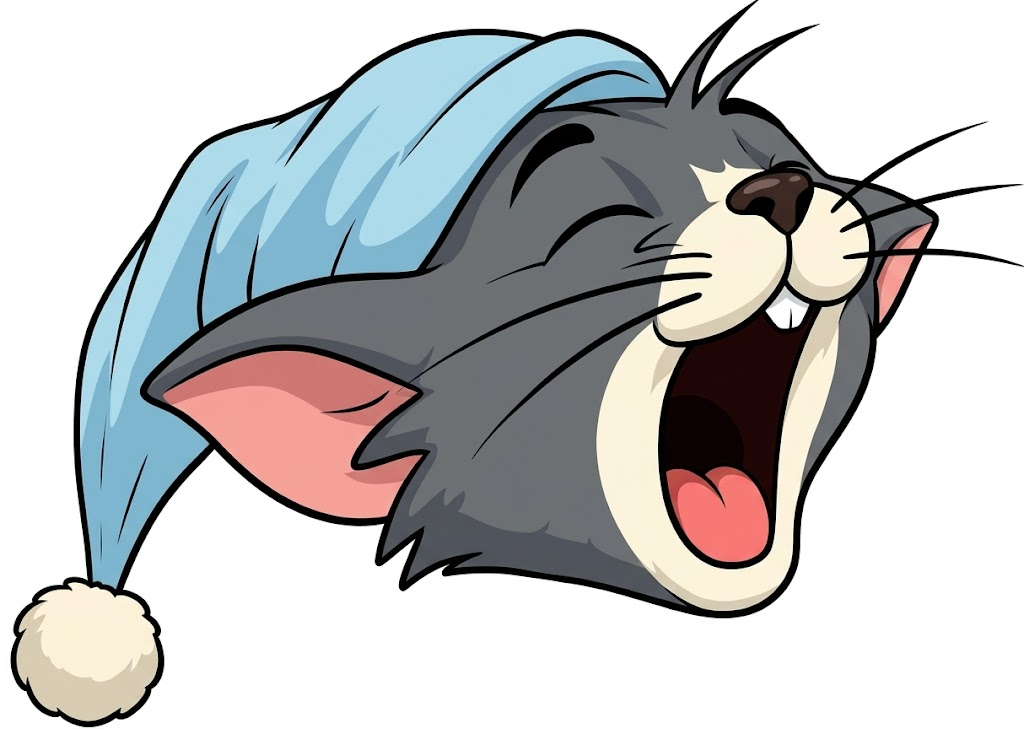}}: Early Token Compression Completes Fast Video Understanding}

\author{
    Hesong Wang$^{1,2,3,\star}$,
    Xin Jin$^{2,\star}$,
    Lu Lu$^{3,\dagger}$,
    Chenhaowen Li$^{3}$, 
    Jian Chen$^{3}$, 
    Qiang Liu$^{3}$, 
    Huan Wang$^{2,\dagger}$\\
    $^1$Zhejiang University \enspace $^2$Westlake University \enspace $^3$Alibaba Cloud Computing \\
    \tt\small \{wanghesong, jinxin86, wanghuan\}@westlake.edu.cn \enspace  ll200214@alibaba-inc.com \\
    \url{https://viridisgreen.github.io/EarlyTom}
}

\begin{document}

\twocolumn[{
    \maketitle
    \begin{figure}[H]
    \hsize=\textwidth
    \vspace{-3.5em}
    \centering
    \includegraphics[width=0.9\textwidth]{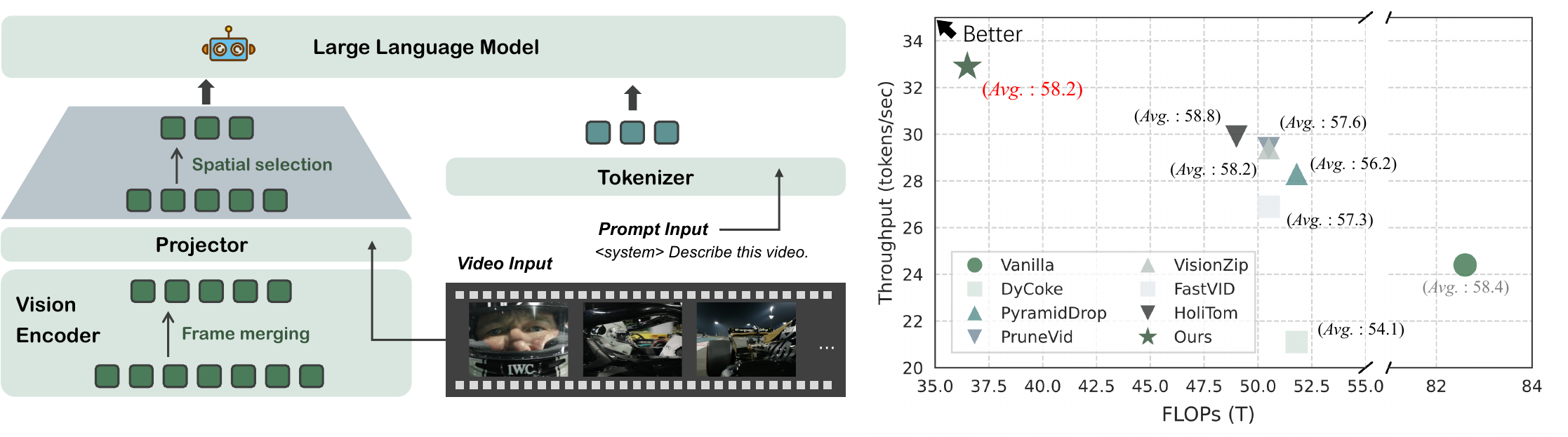}
    \vspace{-1.em}
    \captionof{figure}{
        \textbf{Left}: This paper aims to improve the inference efficiency of video understanding based on video large language models (LLMs). Latency profiling suggests the major speed bottleneck lies in the vision encoder part instead of the LLM. Knowing this, we introduce \textit{EarlyTom}, a training-free \underline{to}ken co\underline{m}pression method designed for the \underline{early} stage (\textit{i.e.}, vision encoder) of video LLMs. EarlyTom features two core components: (1) early-stage visual token compression achieved via inner vision encoder frame merging, and (2) a spatial token selection strategy that further increases compression effectiveness without introducing bias. \textbf{Right}: Scatter plot illustrating the relationship between FLOPs and throughput, along with the average performance across four widely used video understanding benchmarks (MVBench, EgoSchema, LongVideoBench, and VideoMME) for several training-free state-of-the-art methods. EarlyTom achieves state-of-the-art performance while maintaining accuracy comparable to full-token methods.
    }
    \label{fig:teaser}
    \end{figure}
}]

\begin{NoHyper}
\def\thefootnote{}\footnotetext{$^\star$Equal contribution. $^\dagger$Corresponding author. Work done while Hesong Wang was an intern at Alibaba Cloud Computing.}
\end{NoHyper}
\def\thefootnote{\arabic{footnote}}

\begin{abstract}
Video large language models (Video-LLMs) have demonstrated strong capabilities in video understanding tasks. However, their practical deployment is still hindered by the inefficiency introduced by processing massive amounts of visual tokens. Although recent approaches achieve extremely low token retention ratios while maintaining accuracy comparable to full-token baselines, most of them perform compression only at the late stage of prefilling, leaving the efficiency of the vision encoder unoptimized. In this paper, we first show that vision encoding contributes a large portion to the time-to-first-token (TTFT). Therefore, instead of compressing visual tokens only after the vision encoder, performing compression inside the encoder still leaves substantial room for exploration. Based on this insight, we propose EarlyTom, a training-free token compression framework that performs early-stage visual token compression inside the vision encoder, enabling significantly better TTFT reduction and higher throughput. In addition, we introduce a decoupled spatial token selection strategy that improves the overall compression effectiveness. EarlyTom reduces TTFT by up to 2.65$\times$ and FLOPs by up to 61\% on a single NVIDIA A100 GPU for the LLaVA-OneVision-7B model, while maintaining accuracy comparable to the full-token baseline. These improvements substantially enhance the practicality of deploying Video-LLMs in real-world production scenarios.
\vspace{-1.0 em}
\end{abstract}

\section{Introduction}
\label{sec:intro}

\begin{figure}[!t]
    \centering
    \includegraphics[width=1.0\linewidth]{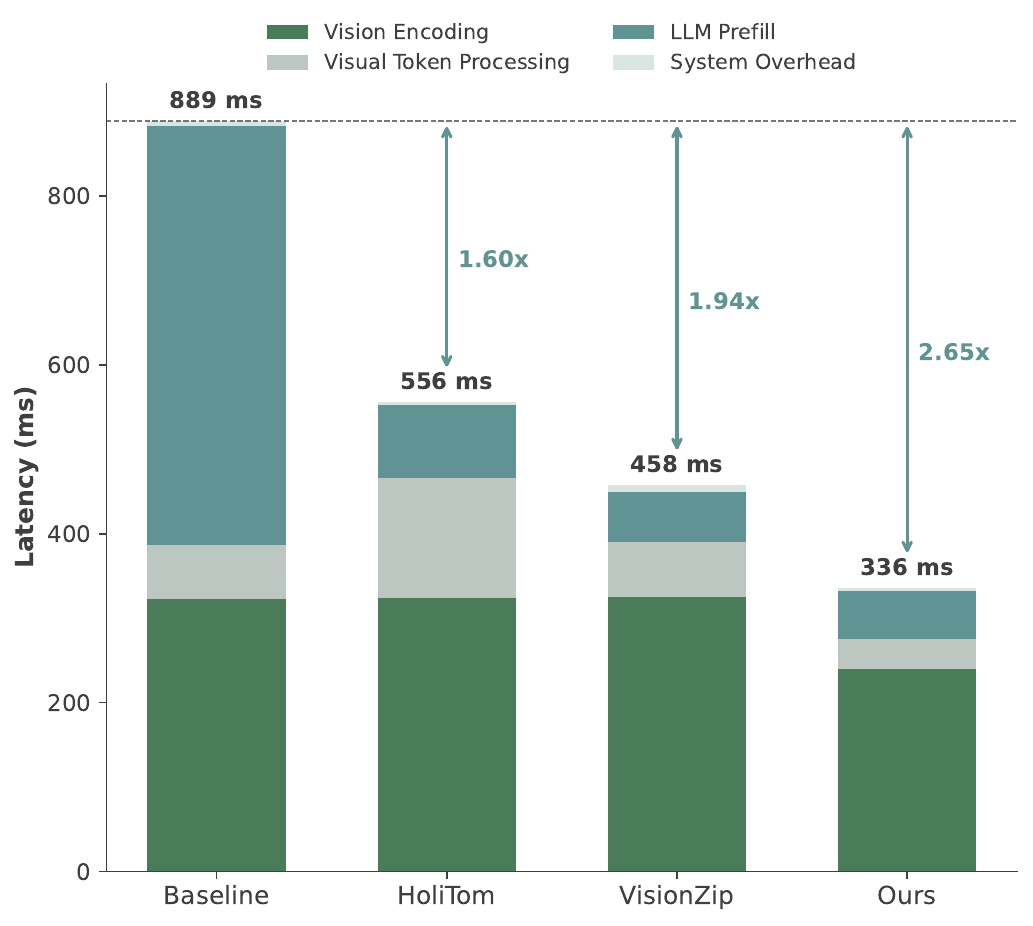}
    \caption{\textbf{Time-to-first-token (TTFT) latency composition.} We break down TTFT into four parts: vision encoding, visual token processing, LLM prefill, and system overhead. In the baseline, vision encoding takes 323 ms, accounting for 36.3\% of the total, indicating that this stage still has substantial room for optimization. For state-of-the-art methods like HoliTom and VisionZip, vision encoding remains the largest component, occupying 55.8\% (324 ms) and 68.4\% (325 ms), respectively. In addition, HoliTom introduces extra token-processing overhead, increasing this component by 121.9\% (+78 ms) compared to the baseline. In contrast, our method reduces vision encoding time directly inside the encoder, achieving a 2.65$\times$ TTFT reduction over the baseline while adding almost no additional overhead, evaluated under 10\% token retention on an NVIDIA A100 GPU.
    }
    \label{fig:ttft-analysis}
    \vspace{-6.5mm}
\end{figure}

Video large language models (Video-LLMs)~\citep{li2025llavaonevision, zhang2024llavavideo, zhang2025videollama, bai2025qwen2-vl, chen2024internvl, wu2024vila, yang2025cambrian-s, li2024llama-vid, li2023videochat, maaz2024video} have demonstrated impressive capability in video understanding tasks. However, efficiently processing large volumes of visual tokens is computationally expensive, which significantly limits the practical deployment of Video-LLMs in real-world scenarios. Although existing methods have made notable progress in compressing vision tokens to improve efficiency, most of them overlook the vision encoder itself. As illustrated in Figure~\ref{fig:ttft-analysis}, the vision encoding stage consumes 36.3\% of the total time-to-first-token (TTFT) in the baseline, and this issue becomes even more pronounced in state-of-the-art methods such as HoliTom and VisionZip, where it rises to 55.8\% and 68.4\%, respectively. As a result, there is still large room to improve the performance of Video-LLMs.

As summarized in prior works~\citep{shao2025holitom, tao2025dycoke, shao2025tokens}, most existing token compression methods operate either after the vision encoder or inside the LLM. Inner-LLM token compression methods, such as FastV~\citep{chen2024fastv}, SparseVLM~\citep{zhang2025sparsevlm}, and PyramidDrop~\citep{xing2025pyramiddrop}, focus on compressing tokens within the LLM and therefore provide limited reduction in TTFT. On the other hand, outer-LLM strategies (e.g., VisionZip~\citep{yang2025visionzip} and LLaVAPruMerge~\citep{shang2025llavaprume}) compress tokens before entering the LLM, offering higher but still limited TTFT reduction. Hybrid approaches such as HoliTom~\cite{shao2025holitom}, FastVID~\citep{shen2025fastvid}, and DyCoke~\citep{tao2025dycoke} attempt to combine both paradigms but still face constrained acceleration, which fundamentally restricts their practicality in compute-bound applications like large-scale video retrieval. Addressing TTFT bottlenecks in video LLMs remains an open challenge.

To better understand the problem, we profile the TTFT composition across several state-of-the-art methods. The results in Figure~\ref{fig:ttft-analysis} reveal that vision encoding accounts for a major portion of TTFT, especially in methods already optimized for LLM prefill latency. In addition, existing compression methods introduce non-trivial computational overhead, which further increases TTFT. These observations motivate us to design a token compression mechanism that acts early inside the vision encoder while minimizing extra overhead for faster and efficient inference.

In this paper, we present \textit{EarlyTom}, an efficient token compression framework designed for extreme performance. Specifically, we propose (1) an inner vision encoder frame merge strategy that compresses redundant visual information during the encoding process, and (2) a decoupled token selection strategy co-designed at the system level to further reduce visual tokens with minimal latency. On LLaVA-OneVision-7B, with only 10\% token retention, EarlyTom achieves 2.65$\times$ TTFT reduction and 1.3$\times$ throughput speedup, while maintaining competitive downstream quality across diverse video understanding benchmarks.

Our main contributions are summarized as follows:
\begin{enumerate}[(a)]
\setlength\topsep{0.0em}
\setlength\itemsep{0.10em}
\setlength\leftmargin{0.5em}
    \item We propose an inner vision encoder frame merge mechanism that compresses redundant visual information during vision encoding, effectively reducing visual tokens with negligible overhead and significantly reducing time-to-first-token.

    \item We introduce a decoupled token selection strategy that performs efficient, low-latency token reduction, further shrinking vision tokens and enabling substantial end-to-end acceleration without sacrificing accuracy.

    \item Extensive experiments on LLaVA-OneVision-0.5B/7B demonstrate that EarlyTom achieves state-of-the-art acceleration performance, delivering extremely fast TTFT while maintaining comparable accuracy.
\end{enumerate}

\section{Related Work}
\label{sec:related_work}

\noindent \textbf{Intra-encoder token compression.} Intra-encoder methods perform token compression within the vision encoder or projector, before tokens are fed into the language model. ToMe~\citep{bolya2023token} reduces tokens in the vision encoder depending on the similarity of key tokens, which improves efficiency and acceleration. PiToMe~\citep{tran2024pitome} proposes an energy score to preserve informative tokens; large similar clusters are merged, while unique tokens with low energy are retained. LLaVAPruMerge~\citep{shang2025llavaprume} selects cluster centers based on attention scores from the \texttt{[CLS]} tagged tokens, then merges the remaining tokens with lower attention scores through KNN clustering~\citep{guo2003knn} and a weighted cluster center update mechanism. VisionZip~\citep{yang2025visionzip} retains visual tokens with higher attention scores, then merges the remaining tokens through clustering. FiCoCo~\citep{han2026filter} integrates multi-dimensional redundant evaluations, token-adaptive association matching, and weighted fusion strategies through a ``filtering-association-compression" process. MustDrop~\citep{liu2024mustdrop} proposes merging similar neighborhood tokens while retaining key tokens in the visual encoder, and by employing dual attention filtering during the prefilling stage to eliminate text-irrelevant tokens. TokenPacker~\citep{li2025tokenpacker} introduces an efficient visual projector with a coarse-to-fine design: it first generates low-resolution point queries via bilinear interpolation, then refines them by injecting high-resolution multi-level visual features through a region-to-point module. MergeMix~\citep{jin2025mergemix} proposes a preference tuning by building augmented samples and training with token merge for efficiency.

\begin{figure*}[!t]
    \centering
    \includegraphics[width=1.0\linewidth]{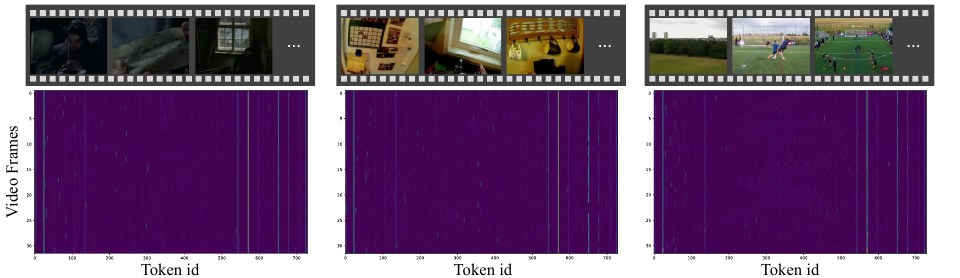}
    \caption{\textbf{The video sink tokens.} We visualize videos across datasets to illustrate the video attention sinking phenomenon: certain tokens (specific frames/regions) consistently attract disproportionately high attention (as shown in the attention score heatmaps), revealing that existing top-K-based token compression methods overlook semantic information in other frames and limit video context understanding.}
    \label{fig:video_sink}
\end{figure*}

\vspace{0.5em}
\noindent \textbf{Pre-LLM token compression.} Pre-LLM methods perform token compression before the language model and after the vision encoder, treating the compression as a plug-and-play module. DyCoke~\citep{tao2025dycoke} proposes a training-free two-stage compression pipeline that merges redundant frame tokens through cross-frame temporal compression, followed by dynamic KV cache pruning during decoding to eliminate spatial redundancy while dynamically preserving key tokens. FastVID~\citep{shen2025fastvid} analyzes video redundancy from temporal and visual density perspectives, proposing dynamic temporal segmentation and density-driven spatio-temporal pruning. It segments videos and prunes based on local ``information density". PVC~\citep{yang2025pvc} proposes a training strategy that progressively encodes each frame and adaptively compresses redundant tokens by leveraging temporal redundancy. VScan~\citep{zhang2025vscan} conducts systematic empirical research on how LLM handles visual tokens, merging them during visual encoding and introducing fine-grained pruning at intermediate model layers. HoliTom~\citep{shao2025holitom} emphasizes global and redundancy-aware holistic compression, reducing tokens by outer-LLM spatio-temporal segmentation and merging while incorporating a robust inner-LLM merging strategy. QueCC~\citep{li2024qucc} analyzes the trade-off between visual tokens and LLM size via inference-time scaling laws, showing that under fixed compute, visual reasoning favors larger LLMs with aggressive token compression, and proposes a query-aware method for extreme compression.

\begin{figure*}
    \centering
    \includegraphics[width=1.0\linewidth]{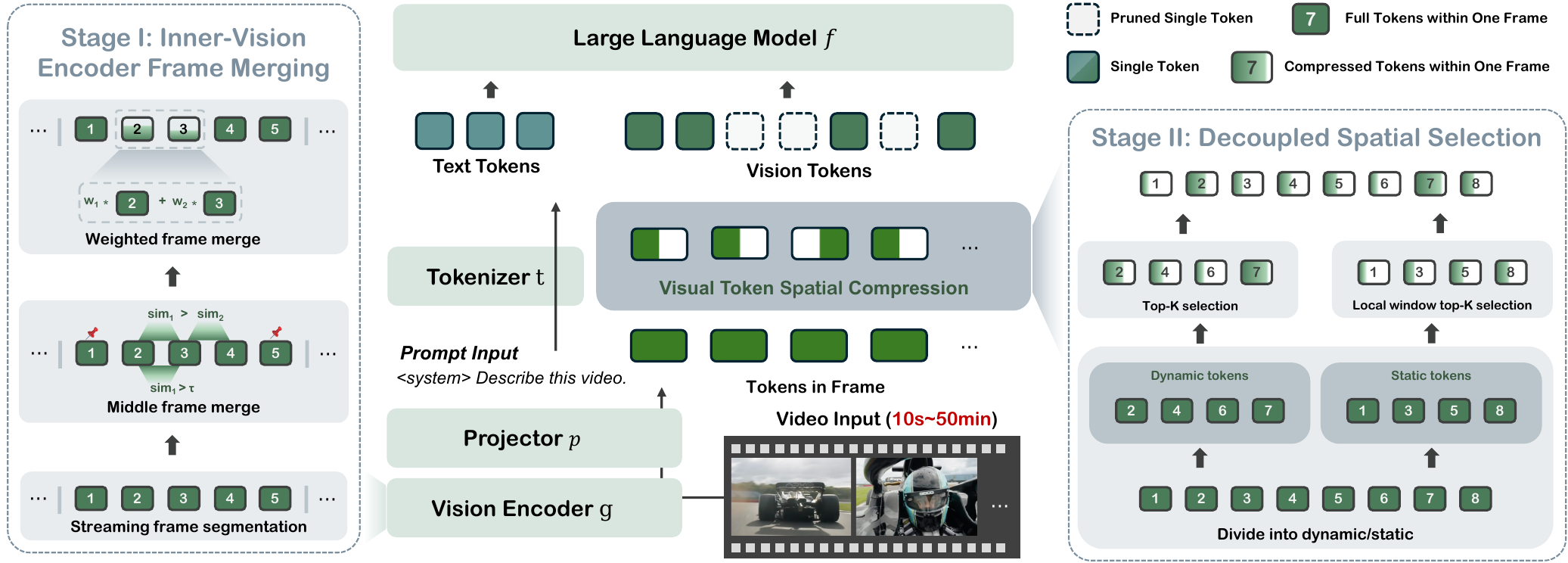}
    \caption{\textbf{Overall pipeline of EarlyTom.}
    Our method consists of two main stages for efficient video token compression. \textbf{Stage I}: Inner-vision encoder frame merging performs temporal compression inside the vision encoder. The video is adaptively segmented based on streaming frame similarity, redundant middle frames are merged using a local-optimal criterion, and merged representations are further refined with weighted fusion to reduce early-stage temporal redundancy. \textbf{Stage II}: Decoupling selection conducts spatial token reduction after vision encoding. Merged frame features are decomposed into dynamic and static token sets: dynamic frames undergo global Top-K selection, while static frames use local-window selection to preserve spatial distribution. The selected tokens from both paths are recombined and fed into the LLM for decoding. Together, these two stages enable early temporal compression and balanced spatial sampling, significantly accelerating Video LLM inference while maintaining semantic fidelity.}
    \label{fig:pipeline}
\end{figure*}

\section{Method}
\label{sec:method}

In this section, we present EarlyTom, a training-free token compression framework for efficient video LLM inference. The overall pipeline is illustrated in Figure~\ref{fig:pipeline} and detailed in the following sections.

\subsection{Preliminaries and Analysis}
\label{sec:3.1}

\noindent \textbf{Video-LLM inference.} The inference process of video LLMs can be divided into three main stages: vision encoding, LLM prefilling, and decoding. During vision encoding, video frames are transformed into embedding representations, which are then aligned to the LLM embedding space through a projector to form video tokens. These video tokens are subsequently concatenated with text tokens and fed into the LLM during the prefilling stage. Finally, the LLM generates responses in an autoregressive manner during decoding. Our method primarily focuses on optimizing the vision encoding and pre-prefilling stages to reduce latency while preserving accuracy.

\vspace{0.5em}
\noindent \textbf{Profiling of time-to-first-token.}  To identify the primary bottlenecks in video LLM inference, we decompose the time-to-first-token latency into four components: vision encoding, visual token processing, LLM prefill, and system overhead. As illustrated in Figure~\ref{fig:ttft-analysis}, vision encoding occupies a substantial portion of TTFT. In the baseline setting, vision encoding accounts for 36.3\% of the total TTFT, and this proportion becomes even more pronounced when applying LLM-prefill–optimized methods such as HoliTom and VisionZip, where it rises to 55.8\% and 68.4\%, respectively. Meanwhile, HoliTom introduces additional compression overhead during the visual token processing stage, further increasing the first-token latency.

\vspace{0.5em}
\noindent \textbf{Video sink tokens.} To analyze how visual tokens contribute to cross-frame information, we visualize SigLIP~\citep{zhai2023siglip} attention maps across video frames. We find that certain spatial patch locations consistently receive unusually high attention, forming vertical stripes across frames even when visual content changes. Some works~\citep{xiao2025efficient, darcet2024vitneedreg, kang2025see, gu2025sink-survey, feng2026edit, zhao2025g-search, zuhri2025softpick, zhuang2025st3} have shown that these correspond to sink tokens, whose query/key vectors exhibit abnormally large norms. Formally, for attention $\text{A}(i,j)=\frac{Q_i K_j^\top}{\sqrt{d}}$, sink tokens satisfy $|Q_{\text{sink}}|_2 \gg |Q_p|_2$, forcing $\text{A}(\text{sink}, j)$ to dominate regardless of content. Thus, raw attention scores from SigLIP cannot directly indicate token importance, since a portion of attention is absorbed by these structural attractors rather than meaningful visual regions.

Based on the above analysis, we propose EarlyTom, which consists of two core components: (1) an inner–vision encoder frame compression stage that improves prefill efficiency with minimal overhead, and (2) a decoupled spatial token selection stage that provides additional token compression without introducing bias into the visual features.

\subsection{Inner Vision Encoder Frame Compression}
\label{sec:3.2}

As analyzed in Section~\ref{sec:3.1}, compressing redundant frames within the vision encoder, which is in the early prefill stage, is crucial for further enhancing model efficiency and performance. Based on this observation, we propose an inner vision encoder frame merge strategy.

\begin{figure}
    \centering
    \includegraphics[width=1.0\linewidth]{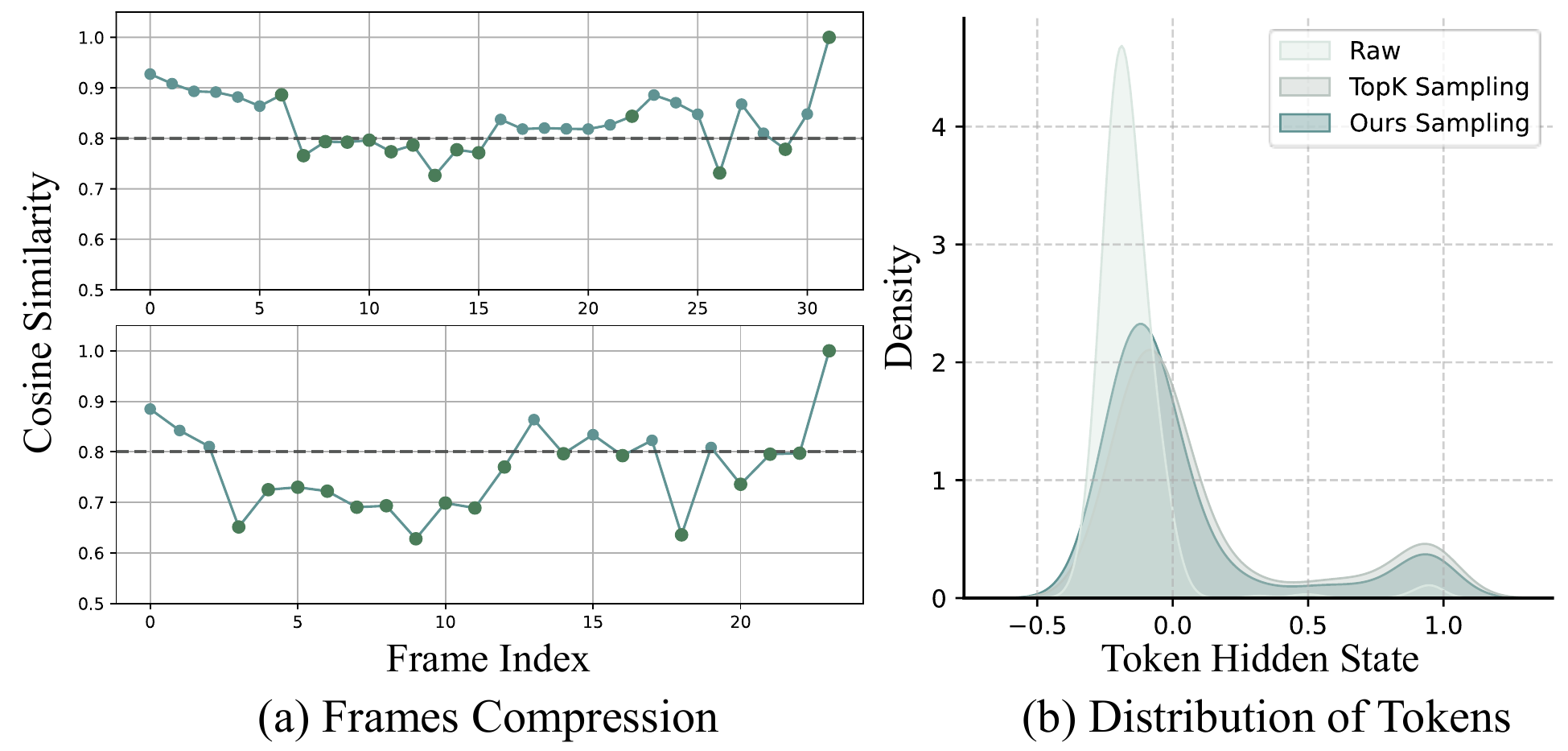}
    \caption{\textbf{Frames compression and distribution of features.} (a) Illustrates the cosine similarity changes across different frame indices for network layers at indices 6 and 20 during frame compression in the vision encoder. (b) The distribution of raw tokens, top-K sampling, and our method. This subfigure shows that our method is closer to vanilla top-K selection.}
    \label{fig:frame-cp-and-distribution}
\end{figure}

\vspace{0.5em}
\noindent \textbf{Streaming frame segmentation.} Given an input video, we perform frame merging at several selected layers in the vision encoder as illustrated in Figure~\ref{fig:frame-cp-and-distribution}. Specifically, we first divide the video into segments according to frame similarity in a streaming manner, which is computed by averaging the cosine similarities of tokens at corresponding spatial positions. For two consecutive frames, we calculate their cosine similarity and update the score with an Exponential Moving Average (EMA) over time. When the similarity score drops below a predefined threshold, we treat this point as a segment boundary, which is described in the equation below:
\begin{equation}
    \begin{aligned}
    \hat{s}_t = \alpha s_t + (1 - \alpha)\hat{s}_{t-1}, \
    \text{break if } \hat{s}_t < \tau_{\mathrm{seg},}
    \end{aligned}
    \label{eq:ema_seg}
\end{equation}
where $\alpha$ denotes the EMA smoothing factor, $s_t$ denotes the cosine similarity between frame $t$ and $t-1$, and $\hat{s}_t$ is the EMA-smoothed similarity. We split the two frames when the $\hat{s}_t$ is smaller than the threshold $\tau_{\mathrm{seg}}$.

\vspace{0.5em}
\noindent \textbf{Middle frame merge.} We adopt a local optimal strategy for the middle frames (i.e., frames within a segment excluding the first and last frames). Two frames are merged if and only if (1) their similarity is higher than a predefined threshold and (2) this similarity is larger than that between the next pair of frames. This process is defined as: 
\begin{equation}
    \begin{aligned}
    \mathrm{merge}(F_i,F_{i+1}) \quad \mathrm{iff} \quad
        \begin{cases}
            s_i>\tau_{\mathrm{merge}} \\
            s_i>s_{i+1}
        \end{cases},
    \end{aligned}
    \label{eq:merge_condition}
\end{equation}
where $s_i$ is the similarity between $F_i$ and $F_{i+1}$, and $\tau_{\mathrm{merge}}$ is the merging threshold.
This merging strategy ensures that only the most similar frames are merged, helping remove redundancy while keeping temporal consistency.

\vspace{0.5em}
\noindent \textbf{Weighted frame merge.} To further improve the quality of merged representations, we use a weighted merging scheme as illustrated in the equation below: 
\begin{equation}
    \hat{F}=\frac{s_{i}F_i+s_{i+1}F_{i+1}}{s_i+s_{i+1}}
    \label{eq:weighted_merge},
\end{equation}
where $F_i$ and $F_{i+1}$ are the frame features and $s_i$, $s_{i+1}$ are their corresponding similarity scores. Each pair of frames is weighted by its similarity with the following frame. This weighting makes the merged frame representation more concentrated around semantically important content and reduces ambiguity caused by uneven temporal variation.

\subsection{Decoupled Spatial Token Selection}
\label{sec:3.3}

In video feature tokens, we observe that certain vision sink tokens, as illustrated in Figure~\ref{fig:video_sink}, consistently appear across all frames, receive high attention scores, and occupy the same positions along the sequence length. Existing methods, such as FastVID~\citep{shen2025fastvid} and HoliTom~\citep{shao2025holitom}, employ Top-K sampling for spatial token merging, which may introduce inherent bias and cause significant distribution shifts across frames as shown in Figure~\ref{fig:frame-cp-and-distribution}. To address this issue, we propose a decoupled sampling strategy that divides all frames into dynamic and static parts and applies distinct sampling schemes for each. Moreover, we adopt a system co-design approach to further enhance efficiency.

\vspace{0.5em}
\noindent \textbf{Decoupling frames into dynamic and static.} 
After merging frames in the vision encoder, we first divide the merged video frames $\hat{F} \in \mathbb{R}^{N \times L \times D}$ into a dynamic part $\hat{F}^d \in \mathbb{R}^{T \times L \times D}$ and a static part $\hat{F}^s \in \mathbb{R}^{(N - T) \times L \times D}$. The division strategy is similar to the streaming segmentation described in Section~\ref{sec:3.2}: we designate the head and tail frames within each segment as dynamic frames, while treating the middle frames as static frames, as we empirically observe that head and tail frames possess the highest discriminative power per segment. Next, we independently compress the dynamic and static frames using their respective strategies.

\vspace{0.5em}
\noindent \textbf{Global top-K selection.} For each dynamic frame, we perform a global Top-K selection based on its per-token attention scores. This process is defined as:
\begin{equation}
    \begin{aligned}
        \hat{\hat{F}}^d_i = \hat{F}^d_i[I_i, :], \
        I_i = \text{TopK}(A_i, \hat{r}), \
        i \in [1, T],
    \end{aligned}
    \label{eq:dynamic_index}
\end{equation}
where $A_i$ denotes the per-token attention scores of frame $F_i$, $I_i$ represents the indices of the selected tokens, and $\hat{r}$ is the re-scaled selection ratio used to achieve the predefined compression rate, incorporated with stage 1, defined as:
\begin{equation}
    \begin{aligned}
        \hat{r} = \frac{r}{(\frac{B-N}{B})*L}, 
    \end{aligned}
    \label{eq:r_hat}
\end{equation}
where $B$ is the number of initial frames (e.g., 32 for LLaVA-OneVision). By performing global importance-based compression, this process further improves the compression ratio while preserving the most motion-sensitive tokens across the entire temporal dimension.

\vspace{0.5em}
\noindent \textbf{Local window top-K selection.} For static frames, our goal is to compress them while preserving their original distribution as much as possible, thereby avoiding unnecessary bias introduced by sink tokens. To this end, we apply a local-window Top-K selection strategy to the static frames. We first divide them into $M$ local windows of equal size:
\begin{equation}
    \begin{aligned}
    \{ W_1, W_2, \dots, W_m \}, \quad
    M = \Big\lceil \frac{L}{w} \Big\rceil, \quad
    w = \Big\lfloor \frac{L}{\hat{r}} \Big\rfloor.
    \end{aligned}
\label{eq:local_window}
\end{equation}
Within each window $W_i$, we select the token with the maximum attention score, finally, we observe compressed static frames $\hat{\hat{F}}^s$. With this technique, the compressed static frames exhibit a distribution that is closer to the original one, thereby mitigating the negative effects caused by the bias introduced by vision sinks.

For all dynamic frames $\hat{\hat{F}}^d$ and static frames $\hat{\hat{F}}^s$, we concatenate them according to their initial order: 
\begin{equation}
    \begin{aligned}
    \hat{\hat{F}} = \text{Gather}(\hat{\hat{F}}^d, \hat{\hat{F}}^s),
    \end{aligned}
    \label{eq:gather}
\end{equation}
which serves as the input for LLM decoding.

\vspace{0.5em}
\noindent \textbf{System co-design.} To further improve execution efficiency, we offload part of the static token selection to the CPU. We empirically observe that dynamic token selection is more time-consuming due to its larger candidate set. As described in Section~\ref{sec:3.1}, all frames are first divided into similarity-based segments; accordingly, we perform segment-wise static token selection on the CPU, while the GPU determines which dynamic tokens should be preserved. With this CPU–GPU heterogeneous computation, we further leverage otherwise idle CPU computational capacity, thereby increasing processing speed while maintaining overall cost-efficiency.

\begin{table*}[t]
    \centering
    \caption{\textbf{Performance and accuracy comparison with SoTA methods across benchmarks.} \textbf{Best} results are in bold, \underline{second-best} results are underlined. Time-to-first-token is denoted as TTFT for simplicity. All efficiency results are measured on a single NVIDIA A100 GPU.}
    \setlength{\tabcolsep}{0.8mm}
    \resizebox{1.0\linewidth}{!}{
        \begin{tabular}{l|ccccc|cccc|cc}
        \toprule
        \multirow{2}{*}{Method} &
        \multirow{2}{*}{\shortstack{Before LLM\\Retained Ratio}} &
        \multirow{2}{*}{\shortstack{Prefilling\\FLOPs (T) $\downarrow$}} &
        \multirow{2}{*}{\shortstack{FLOPs\\Ratio $\downarrow$}} &
        \multirow{2}{*}{\shortstack{TTFT\\(ms) $\downarrow$}} &
        \multirow{2}{*}{\shortstack{Throughput\\(tokens/s) $\uparrow$}} &
        \multirow{2}{*}{\shortstack{MVBench\\$\uparrow$}} &
        \multirow{2}{*}{\shortstack{EgoSchema\\$\uparrow$}} &
        \multirow{2}{*}{\shortstack{LongVideo\\Bench $\uparrow$}} &
        \multirow{2}{*}{\shortstack{VideoMME\\$\uparrow$}} &
        \multicolumn{2}{c}{Avg. $\uparrow$} \\
         & & & & & & & & & & Score & \% \\
            \midrule
            \rowcolor{gray!20} 
            LLaVA-OV-7B & 100\% & 82.6 & 100\% & 889.9 & 24.4 & 58.3 & 60.4 & 56.4 & 58.6 & 58.4 & 100 \\ %
            FastV~\citep{chen2024fastv}$_{\text{ECCV'24}}$       & 100\% & 51.1 & 61.9\% & 820.0 & 28.4 & 55.9 & 57.5 & 56.7 & 56.1 & 56.5 & 96.7 \\ %
            PyramidDrop~\citep{xing2025pyramiddrop}$_{\text{CVPR'25}}$  & 100\% & 51.8 & 62.7\% & 813.4 & 28.3 & 56.1 & 58.0 & 54.1 & 56.4 & 56.2 & 96.2 \\
            DyCoke~\citep{tao2025dycoke}$_{\text{CVPR'25}}$      & 25\% & 50.5 & 61.1\%  & 905.6 & 21.1 & 53.1 & 59.5 & 49.5 & 54.3 & 54.1 & 92.6 \\
            VisionZip~\citep{yang2025visionzip}$_{\text{CVPR'25}}$    & 25\% & 50.5 & 61.1\%  & 516.6 & 29.4 & 57.9 & 60.3 & 56.5  & 58.2 & 58.2 & 99.7 \\ %
            PruneVid~\citep{huang2025prunevid}$_{\text{ACL'25}}$    & 25\% & 50.5 & 61.1\%  & 703.6 & 29.4 & 57.4 & 59.9 & 55.7 & 57.4 & 57.6 & 98.6 \\ %
            FastVID~\citep{shen2025fastvid}$_{\text{NeurIPS'25}}$    & 25\% & 50.5 & 61.1\% & \underline{581.6} & 26.9 & 56.5 & 58.2 & 56.3 & 58.0 & 57.3 & 98.1 \\
            HoliTom~\citep{shao2025holitom}$_{\text{NeurIPS'25}}$     & 25\% & \underline{49.0} & \underline{59.3\%} & 661.3 & \underline{29.9} & 58.4  & 61.2 & 56.7 & 58.9 & 58.8 & 100.7 \\
            \rowcolor{lightblue} 
            \textbf{EarlyTom}    & 25\% & \textbf{36.5} & \textbf{44.2\%} & \textbf{426.3} & \textbf{32.9} & 57.4  & 60.5 & 56.3 & 58.5 & 58.2 & 99.7 \\
            \midrule
            VisionZip~\citep{yang2025visionzip}$_{\text{CVPR'25}}$    & 20\% & 48.7 & 58.9\% & \underline{495.0} & 29.8 & 57.7 & 59.8 & 55.2 & 57.9 & 57.7 & 98.8 \\ %
            PruneVid~\citep{huang2025prunevid}$_{\text{ACL'25}}$     & 20\% & 49.0 & 59.3\% & 662.1 & 29.5 & 57.2 & 59.7 & 54.7 & 56.9 & 57.1 & 97.8 \\ %
            FastVID~\citep{shen2025fastvid}$_{\text{NeurIPS'25}}$    & 20\% & 48.7 & 58.9\% & 546.6 & 27.6 & 56.3 & 57.9 & 57.1  & 57.9 & 57.3 & 98.1 \\
            HoliTom~\citep{shao2025holitom}$_{\text{NeurIPS'25}}$    & 20\% & \underline{47.5} & \underline{57.5\%} & 622.3 & \underline{30.0} & 58.7  & 61.0  & 57.1  & 58.6  & 58.8  & 100.7  \\ %
            \rowcolor{lightblue} 
            \textbf{EarlyTom}    & 20\% & \textbf{35.1} & \textbf{42.4\%} & \textbf{415.3} & \textbf{33.4} & 57.8  & 60.6 & 55.6 & 58.0 & 58.1 & 99.3 \\
            \midrule
            VisionZip~\citep{yang2025visionzip}$_{\text{CVPR'25}}$   & 15\% & 46.9 & \underline{56.8\%} & \underline{475.9} & \textbf{32.1} & 56.5 & 59.8 & 54.4 & 56.1 & 56.7 & 97.1 \\ %
            PruneVid~\citep{huang2025prunevid}$_{\text{ACL'25}}$    & 15\% & 47.5 & 57.5\% & 574.1 & 27.1 & 56.8 & 59.7 & 55.4 & 56.6 & 57.1 & 97.8 \\ %
            FastVID~\citep{shen2025fastvid}$_{\text{NeurIPS'25}}$    & 15\%  & 46.9 & 56.8\% & 530.8 & 28.7 & 56.0 & 57.4 & 56.2 & 57.7 & 56.8 & 97.3 \\
            HoliTom~\citep{shao2025holitom}$_{\text{NeurIPS'25}}$    & 15\% & \underline{46.0} & 55.7\% & 572.7 & 27.5 & 58.1  & 61.2  &  56.4  &  57.3  &  58.2  &  99.7  \\
            \rowcolor{lightblue} 
            \textbf{EarlyTom}    & 15\% & \textbf{33.6} & \textbf{40.7\%} & \textbf{390.6} & \underline{30.4} & 57.5  & 60.2 & 54.4 & 56.9 & 57.3 & 98.1 \\
            \midrule
            VisionZip~\citep{yang2025visionzip}$_{\text{CVPR'25}}$   & 10\%  & 45.2 & 54.7\% & \underline{458.5} & 28.5 & 53.5 & 58.0 & 49.3 & 53.4 & 53.5 & 91.6 \\ %
            PruneVid~\citep{huang2025prunevid}$_{\text{ACL'25}}$    & 10\% & 45.9 & 55.6\% & 592.2 & 28.6 & 56.2 & 59.8 & 54.5 & 56.0 & 56.6 & 96.9 \\ %
            FastVID~\citep{shen2025fastvid}$_{\text{NeurIPS'25}}$    & 10\%  & 45.2 & 54.7\% & 502.1 & 28.3 & 55.9 & 56.5 & 56.3 & 57.3  & 56.5 & 96.7 \\
            HoliTom~\citep{shao2025holitom}$_{\text{NeurIPS'25}}$    & 10\% & \underline{44.6} & \underline{54.0\%} & 556.6 & \underline{29.0} & 57.3  & 61.2  &  56.3  & 56.8 & 57.9  & 99.1  \\
            \rowcolor{lightblue} 
            \textbf{EarlyTom}    & 10\% & \textbf{32.2} & \textbf{39.0\%} & \textbf{336.2} & \textbf{31.6} & 56.5  & 60.1 & 52.4 & 55.8 & 56.2 & 96.2 \\
            \bottomrule
        \end{tabular}
    }
    \label{tab:benchmark}
\end{table*}

\section{Experiments}
\label{sec:experiments}

\subsection{Settings}
\label{sec:5.1}

\noindent \textbf{Benchmarks and metrics.} 
In our paper, we choose four mainstream video understanding tasks for our evaluation: MVBench~\citep{li2024mvbench}, EgoSchema~\citep{mangalam2023egoschema}, LongVideoBench~\citep{wu2024longvideobench}, and VideoMME~\citep{fu2025videomme}. The videos in these tasks vary in length and scenario difficulty, providing a comprehensive perspective for evaluating the effectiveness and generalization of our method. To evaluate the efficiency of our approach, we report time-to-first-token (TTFT), throughput, and TFLOPs. These metrics capture both the latency and compute efficiency of our method, highlighting its practical benefits for large-scale or long-form video processing.

\vspace{0.5em}
\noindent \textbf{State-of-the-art methods.} 
To evaluate the performance of our method, we compare our method with some mainstream token compression methods in Video-LLMs, i.e., FastV~\citep{chen2024fastv}, PyramidDrop~\citep{xing2025pyramiddrop}, DyCoke~\citep{tao2025dycoke}, VisionZip~\citep{yang2025visionzip}, FastVid~\citep{shen2025fastvid}, PruneVid~\citep{huang2025prunevid}, and HoliTom~\citep{shao2025holitom}. For their accuracy results, we report results from HoliTom.

\vspace{0.5em}
\noindent \textbf{Implementations.}
Our method is implemented based on the LLaVA-OneVision-0.5B/7B model~\cite{li2025llavaonevision}. We incorporate the inner-LLM merging technique from HoliTom~\citep{shao2025holitom} into our framework and develop a custom Triton kernel to ensure computational efficiency. All experiments are conducted on NVIDIA A100 and RTX 4090 GPUs. The reported time-to-first-token (TTFT) is measured using the NVIDIA Nsight Systems profiler. For throughput evaluation, we report the average result of ten inference runs after warm-up. The prefilling FLOPs are computed following the HoliTom~\citep{shao2025holitom} benchmark protocol, which consists of both vision encoding and LLM prefilling FLOPs. In accordance with the official LLaVA-OneVision configuration, 32 video frames are uniformly sampled as visual inputs, and the vision encoder employs a pretrained SigLIP model~\citep{zhai2023siglip}. Detailed configurations for hyperparameter selection are provided in Table~\ref{tab:hyper} in the Appendix.
All benchmark evaluations are performed using the LMMs-Eval framework~\citep{zhang2024lmms, lmms_eval2024}.

\begin{table*}[t]
    \centering
    \caption{\textbf{Cross-backbone comparison on performance and accuracy.} \textbf{Best} results are in bold, \underline{second-best} results are underlined. Time-to-first-token is denoted as TTFT for simplicity. All efficiency results are measured on a single NVIDIA A100 GPU.}
    \setlength{\tabcolsep}{0.8mm}
    \resizebox{1.0\linewidth}{!}{
        \begin{tabular}{l|ccccc|cccc|cc}
        \toprule
        \multirow{2}{*}{Method} &
        \multirow{2}{*}{\shortstack{Before LLM\\Retained Ratio}} &
        \multirow{2}{*}{\shortstack{Prefilling\\FLOPs (T) $\downarrow$}} &
        \multirow{2}{*}{\shortstack{FLOPs\\Ratio $\downarrow$}} &
        \multirow{2}{*}{\shortstack{TTFT\\(ms) $\downarrow$}} &
        \multirow{2}{*}{\shortstack{Throughput\\(tokens/s) $\uparrow$}} &
        \multirow{2}{*}{\shortstack{MVBench\\$\uparrow$}} &
        \multirow{2}{*}{\shortstack{EgoSchema\\$\uparrow$}} &
        \multirow{2}{*}{\shortstack{LongVideo\\Bench $\uparrow$}} &
        \multirow{2}{*}{\shortstack{VideoMME\\$\uparrow$}} &
        \multicolumn{2}{c}{Avg. $\uparrow$} \\
         & & & & & & & & & & Score & \% \\
            \midrule
            \rowcolor{gray!20} 
            LLaVA-OV-0.5B & 100\% & 45.3 & 100\% & 413.7 & 42.7 & 45.5 & 26.8 & 45.8 & 43.7 & 40.5 & 100 \\ %
            FastVID~\citep{shen2025fastvid}$_{\text{NeurIPS'25}}$   & 25\%  & 42.4 & 93.6\% & 409.9 & 25.9 & 44.7 & 25.3 & 44.9 & 42.1 & 39.3 & 97.0 \\
            VisionZip~\citep{yang2025visionzip}$_{\text{CVPR'25}}$    & 25\% & 42.4 & 93.6\% & \underline{368.6} & \underline{41.1} & 45.6 & 27.7 & 45.9 & 42.9 & 40.5 & 100.0 \\
            HoliTom~\citep{shao2025holitom}$_{\text{NeurIPS'25}}$     & 25\% & \underline{42.3} & \underline{93.4\%} & 519.4 & 35.2 & 45.8 & 27.6 & 46.2 & 44.4 & 41.0 & 101.2 \\
            \rowcolor{lightblue} 
            \textbf{EarlyTom}    & 25\% & \textbf{29.9} & \textbf{66.0\%} & \textbf{331.5} & \textbf{47.8} & 45.5 & 27.4 & 46.3 & 43.4 & 40.7 & 100.4 \\
            \midrule
            FastVID~\citep{shen2025fastvid}$_{\text{NeurIPS'25}}$   & 20\%  & 42.3 & 92.4\% & 412.6 & 28.8 & 43.8 & 25.7 & 44.3 & 41.6 & 38.9 & 96.0 \\
            VisionZip~\citep{yang2025visionzip}$_{\text{CVPR'25}}$    & 20\% & 42.3 & 93.4\% & \underline{368.5} & \textbf{42.3} & 45.1 & 27.5 & 44.8 & 42.7 & 40.0 & 98.8 \\
            HoliTom~\citep{shao2025holitom}$_{\text{NeurIPS'25}}$    & 20\% & \underline{42.2} & \underline{93.2\%} & 499.4 & 38.3 & 45.5 & 27.7 & 45.9 & 44.1 & 40.8 & 100.7 \\
            \rowcolor{lightblue} 
            \textbf{EarlyTom}    & 20\% & \textbf{29.8} & \textbf{65.8\%} & \textbf{313.1} & \underline{40.6} & 45.2 & 27.5 & 44.7 & 43.7 & 40.3 & 99.5 \\
            \midrule
            FastVID~\citep{shen2025fastvid}$_{\text{NeurIPS'25}}$   & 15\%  & 42.1 & 92.9\% & 411.3 & 29.4 & 43.1 & 25.3 & 44.7 & 40.7 & 38.5 & 95.1\\
            VisionZip~\citep{yang2025visionzip}$_{\text{CVPR'25}}$   & 15\% & \underline{42.1} & \underline{92.9\%} & \underline{367.1} & \textbf{37.8} & 44.6 & 26.9 & 44.9 & 42.3 & 39.7 & 98.0 \\
            HoliTom~\citep{shao2025holitom}$_{\text{NeurIPS'25}}$    & 15\% & 42.1 & 92.9\% & 473.9 & 34.1 & 45.4 & 27.6 & 46.4 & 43.4 & 40.7 & 100.4 \\
            \rowcolor{lightblue} 
            \textbf{EarlyTom}    & 15\% & \textbf{29.7} & \textbf{65.6\%} & \textbf{311.1} & \underline{35.1} & 44.8 & 27.0 & 44.9 & 42.3 & 39.8 & 98.3 \\
            \midrule
            FastVID~\citep{shen2025fastvid}$_{\text{NeurIPS'25}}$   & 10\%  & 42.0 & 92.7\% & 408.5 & 31.9 & 42.7 & 24.7 & 44.2 & 40.7 & 38.1 & 94.1 \\
            VisionZip~\citep{yang2025visionzip}$_{\text{CVPR'25}}$   & 10\%  & \underline{42.0} & \underline{92.7\%} & \underline{366.1} & 38.7 & 43.2 & 25.8 & 42.6 & 40.0 & 37.9 & 93.6 \\
            HoliTom~\citep{shao2025holitom}$_{\text{NeurIPS'25}}$    & 10\% & 42.0 & 92.7\% & 457.1 & \underline{39.6} & 45.0 & 27.3 & 44.5 & 43.3 & 40.0 & 98.8 \\
            \rowcolor{lightblue} 
            \textbf{EarlyTom}    & 10\% & \textbf{29.6} & \textbf{65.3\%} & \textbf{280.1} & \textbf{43.9} & 44.3 & 26.8 & 44.5 & 41.8 & 39.4 & 97.3 \\
            \bottomrule
        \end{tabular}
    }
    \label{tab:05b_benchmark}
    \vspace{-4mm}
\end{table*}

\vspace{0.5em}
\noindent \textbf{FLOPs and throughput.}
In our paper, we evaluate inference performance using FLOPs and throughput. Since both the vision encoder and the LLM decoder are built on Transformer architectures, the computation of FLOPs follows the same formulation. The computational cost mainly comes from the multi-head self-attention (MHA) and the feed-forward network (FFN). Following previous works~\citep{chen2024fastv, tao2025dycoke, xing2025pyramiddrop, shao2025holitom}, the FLOPs for processing $L_i$
 vision tokens at layer $i$ with hidden size $D$ and FFN intermediate size $M$, can be expressed as $4L_i D^2 + 2L^2_i D +2L_iDM$. HoliTom~\citep{shao2025holitom} reports that only about 2\% of FLOPs occur during the decoding stage, and the majority of the computation lies in the prefilling (encoder) stage. However, different from HoliTom, we evaluate not only the LLM decoder but also the effectiveness and efficiency of the vision encoder. Therefore, the FLOPs of the whole inference pipeline are computed according to Equation~\eqref{eq:flops}:
\begin{equation}
    \begin{aligned}
        \text{FLOPs} = & \sum_{i=1}^{T_v} \underbrace{\left(4 L_i D^2+2 L_i^2 D+2 L_i DM\right)}_{\text {Vision Encoder FLOPs per layer }} \\
        + & \sum_{i=1}^{T_t} \underbrace{\left(4 L_i D^2+2 L_i^2 D+2 L_i DM\right)}_{\text {LLM Decoder FLOPs per layer }}.
    \end{aligned}
    \label{eq:flops}
\end{equation}
Compared with some outer-LLM token compression methods, performing token compression early within the vision encoder reduces the number of tokens entering the LLM, thereby significantly decreasing FLOPs and improving inference efficiency. For throughput evaluation, we use the same video input for all methods and measure the total runtime $r$. The throughput is reported as the average generated tokens per second over ten runs (with two warm-up passes): $\text{Throughput} = \text{Avg}(\sum^r_{i=1}\frac{\text{tokens}}{\text{time}})$.

\subsection{Main Results}
\label{sec:4.2}

\noindent \textbf{Performance comparison with state-of-the-art methods.} Table~\ref{tab:benchmark} presents a comprehensive comparison of EarlyTom against a range of state-of-the-art training-free token compression methods, focusing on FLOPs, TTFT, and throughput. As shown in Table~\ref{tab:benchmark}, prior methods such as PyramidDrop~\citep{xing2025pyramiddrop}, VisionZip~\citep{yang2025visionzip}, PruneVid~\citep{huang2025prunevid}, and FastVID~\citep{shen2025fastvid} significantly reduce the FLOPs of the prefill stage. However, these approaches largely rely on late-stage compression and operate after vision encoding, leaving the vision encoder as a dominant bottleneck. As a result, although their retained-token ratios fall to as low as 10–25\%, the corresponding TTFT still ranges from 458 ms to 661 ms, and the throughput fluctuates between 27.5 and 32.1 tokens/s. In contrast, EarlyTom fundamentally shifts the compression point to an early stage inside the vision encoder, thereby optimizing one of the most expensive portions of TTFT. Consequently, EarlyTom achieves the lowest TTFT among all training-free approaches, only 336.2 ms with a 10\% retained-token ratio, outperforming all other compared methods by a clear margin.

Meanwhile, EarlyTom maintains a FLOPs budget of 36.5T under a retention ratio of 25\%, achieving significantly higher efficiency than the full-token baseline (82.6T) and other token compression methods. The results indicate that EarlyTom not only reduces the computational burden but also fundamentally improves system-level efficiency by co-optimizing both vision-encoding and LLM-prefill costs. Even under more aggressive compression ratios, EarlyTom maintains low TTFT and high throughput simultaneously, outperforming methods such as VisionZip~\citep{yang2025visionzip}, PruneVid~\citep{huang2025prunevid}, and FastVID~\citep{shen2025fastvid}. This consistent dominance across multiple retention configurations highlights the superiority of EarlyTom in optimizing early-stage token compression and its ability to deliver balanced improvements in both latency and system efficiency. Overall, EarlyTom sets a new benchmark for inference efficiency in video LLMs, significantly outperforming all existing training-free methods in FLOPs, TTFT, and throughput.

\begin{table}[t]
    \centering
    \caption{
        Frame merging effectiveness varies across different initial compression layers. We report with a compression ratio of 0.2.
    }
    \vspace{-1mm}
    \setlength{\tabcolsep}{1.0mm}
    \resizebox{1.0\linewidth}{!}{
        \begin{tabular}{l|cc|ccc|c}
            \toprule
            \textbf{\#Layer} &
            TTFT $\downarrow$ &
            Throughput $\uparrow$ &
            MVBench $\uparrow$ &
            VideoMME $\uparrow$ &
            EgoSchema $\uparrow$ &
            Avg. $\uparrow$\\
           \midrule
            Layer 4   & \textbf{380.0}  & 31.6  & 57.4 & 57.9 & 60.4 & 58.6 \\
            Layer 6   & 387.1  & \textbf{32.3}  & \textbf{57.8} & \textbf{58.1} & 60.4 & \textbf{58.9} \\
            Layer 8   & 421.1  & 30.7  & 57.5 & 58.0 & 60.4 & 58.6 \\
            Layer 10  & 436.9  & 31.1  & 57.4 & 58.0 & \textbf{60.6} & 58.7 \\
            \bottomrule
        \end{tabular}
    }
    \vspace{-2mm}
    \label{tab:ablation_layer}
\end{table}

\begin{table}[t]
    \centering
    \caption{Ablation study of different token sampling ways. We report the throughput and accuracy of three video tasks. In all results, we set the retain ratio to 0.2.}
    \vspace{-1mm}
    \setlength{\tabcolsep}{0.8mm}
    \resizebox{1.0\linewidth}{!}{
        \begin{tabular}{l|c|ccc|c}
            \toprule
            \textbf{Sampling} & Throughput $\uparrow$ & MVBench $\uparrow$ & VideoMME $\uparrow$ & EgoSchema $\uparrow$ & Avg. $\uparrow$ \\
           \midrule
            Random     & \textbf{35.3} & 57.0 & 56.6 & 59.8 & 57.8 \\
            Top-K     & 31.5 & 57.5 & 57.3 & 60.4 & 58.4 \\
            \textbf{EarlyTom}    & 33.4 & \textbf{57.8} & \textbf{58.1} & \textbf{60.6} & \textbf{58.8} \\
            \bottomrule
        \end{tabular}
    }
    \label{tab:sampling}
    \vspace{-4.mm}
\end{table}

\vspace{0.5em}
\noindent \textbf{Accuracy comparison with state-of-the-art methods.} Although EarlyTom is designed for improving efficiency, it also maintains accuracy comparable to a full-token baseline across multiple benchmarks. Table~\ref{tab:benchmark} shows results across four widely used video understanding benchmarks. Under all configurations, EarlyTom achieves an average accuracy of more than 96\% compared with the full-token baseline, which is competitive with other training-free state-of-the-art methods. Meanwhile, EarlyTom achieves this accuracy while reducing TTFT by up to 2.65$\times$, demonstrating that the substantial efficiency gains do not come at the cost of model performance. In more challenging compression scenarios such as the 15\% and 10\% settings, other methods often show noticeable degradation in benchmark performance. For instance, VisionZip~\citep{yang2025visionzip} suffers a noticeable accuracy drop under aggressive pruning, whereas EarlyTom maintains stable performance, with only a 4\% decrease compared to the full-token output, while VisionZip~\citep{yang2025visionzip} drops by nearly 9\%. This indicates that EarlyTom preserves relevant features more effectively than late-stage pruning strategies. In summary, EarlyTom achieves near-baseline accuracy while significantly outperforming all prior approaches in computational efficiency, proving its practical value for real-world, latency-sensitive deployments.

\vspace{0.5em}
\noindent \textbf{Comparison across different backbones.} To evaluate the robustness and generality of EarlyTom, we apply EarlyTom to a smaller backbone, LLaVA-OV-0.5B, and report results in Table~\ref{tab:05b_benchmark}. Similar to the observations on the 7B model, EarlyTom achieves substantial TTFT and FLOPs reductions and throughput improvements across all compression settings, while maintaining benchmark accuracy within a narrow margin of the full-token baseline across the four benchmarks, demonstrating that early-stage compression generalizes well even under lightweight vision-encoder architectures. This robustness is further evidenced across different retained-token settings: EarlyTom yields consistent improvements in efficiency with stable accuracy regardless of the size of the backbone. These results confirm that EarlyTom is architecture-agnostic and capable of delivering strong acceleration without sacrificing quality.

\subsection{Ablation Studies}
\label{sec:4.3}

\noindent \textbf{Contribution of compression modules.}
As shown in Table~\ref{tab:module}, the temporal merge achieves 98.8\% of the baseline accuracy while retaining approximately 73.9\% of the tokens. The spatial token selection module reaches the same accuracy with a retention rate of 20\%. When both frame merge and spatial selection are jointly applied, our method further improves performance, achieving an accuracy of 58.8, surpassing either individual module.

\begin{table}[t]
    \centering
    \caption{Ablation study on the compression module of our method. The stage-1 retention ratio is averaged due to its sample-dependent behavior, as the redundancy strongly depends on the input sample. }
    \vspace{-1mm}
    \setlength{\tabcolsep}{1.0mm}
    \resizebox{1.0\linewidth}{!}{
        \begin{tabular}{l|c|ccc|c}
            \toprule
            \multirow{2}{*}{\textbf{Method}} &
            \multirow{2}{*}{\shortstack{Retained\\Ratio}} &
            \multirow{2}{*}{MVBench $\uparrow$} & 
            \multirow{2}{*}{VideoMME $\uparrow$} & 
            \multirow{2}{*}{EgoSchema $\uparrow$} &
            \multirow{2}{*}{Avg. $\uparrow$}\\
            & & & & & \\
            \midrule
            \rowcolor{gray!20} 
            Vanilla         & 100\%    & 58.3   & 58.6   & 60.4  & 59.1 \\
            Only stage-1    & 73.9\%   & \textbf{57.9} & 57.0 & 60.3 & 58.4 \\
            Only stage-2    & 20\%  & 57.3 & 57.6 & 60.4 & 58.4 \\
            \textbf{EarlyTom}   & 20\%  & 57.8 & \textbf{58.1} & \textbf{60.6} & \textbf{58.8} \\
            \bottomrule
        \end{tabular}
    }
    \vspace{-4.mm}
    \label{tab:module}
\end{table}

\vspace{0.5em}
\noindent \textbf{Impact of different frame merging layers.}
Table~\ref{tab:ablation_layer} demonstrates that initiating the merging process from layer 4 yields the lowest TTFT, but results in a noticeable accuracy degradation. In contrast, starting from layer 6 achieves the best balance between accuracy and throughput. Since our frame merging primarily depends on the hidden states produced by the vision encoder, this also explains why throughput and TTFT do not scale proportionally.

\vspace{0.5em}
\noindent \textbf{Effectiveness of the proposed local window sampling.}
Table~\ref{tab:sampling} shows that the top-K selection is slower than random sampling because it requires a global ranking over all tokens with complexity $O(N \log K)$, whereas random sampling only generates $K$ indices with complexity $O(K)$. As a result, top-K selection incurs extra computational and memory overhead, while random sampling cannot retain the most informative tokens. Our local window sampling combines the strengths of both methods, achieving a better trade-off between efficiency and accuracy.

\section{Conclusion}
\label{sec:conclusion}
In this paper, we propose \textit{EarlyTom}, a training-free token compression framework for fast Video LLM inference. Benefiting from early-stage frame merging within the vision encoder and a further decoupled spatial token selection strategy, EarlyTom achieves up to a 2.65$\times$ reduction in TTFT and a 61\% reduction in FLOPs, while keeping comparable accuracy to the full-token baseline. These results demonstrate the effectiveness and efficiency of EarlyTom, revealing its strong potential in video understanding tasks and laying a solid foundation for the deployment of Video LLMs in real-world production environments.

\section*{Acknowledgement}
\label{sec:acknow}

This paper is supported by Young Scientists Fund of the National Natural Science Foundation of China (NSFC) (No. 62506305), Zhejiang Leading Innovative and Entrepreneur Team Introduction Program (No. 2024R01007), Key Research and Development Program of Zhejiang Province (No. 2025C01026), Scientific Research Project of Westlake University (No. WU2025WF003). It is also supported by the research funds of the National Talent Program, Hangzhou Municipal Talent Program and Alibaba Innovative Research Program.

{
    \small
    \bibliographystyle{ieeenat_fullname}
    \bibliography{reference}
}

\clearpage
\setcounter{page}{1}
\maketitlesupplementary

\setcounter{section}{0}
\renewcommand{\thesection}{\Alph{section}}

\section*{Overview}

Due to page limitations in the main paper, we present additional quantitative experiments, detailed latency analyses, qualitative visualizations, and implementation details in this supplementary material. The content is organized as follows:

\begin{itemize}
    \item \textbf{Section \ref{sec:exp}} evaluates the generalizability of our method on a different video-LLM architecture. Specifically, we provide extensive efficiency and accuracy results on the LLaVA-Video-7B benchmark to verify the robustness of EarlyTom across different backbones. 
    Furthermore, we extend our evaluation to the Qwen2.5-VL architecture, comparing EarlyTom against two native token reduction baselines. We also conduct fine-grained ablation studies to investigate the individual contribution of each component within our framework on Qwen architecture.
    
    \item \textbf{Section \ref{sec:more_ttft}} presents a fine-grained decomposition of the time-to-first-token (TTFT) latency. We analyze the specific contributions of vision encoding, visual token processing, and LLM prefilling to the total latency on both LLaVA-OneVision-7B and 0.5B models across different settings.
    
    \item \textbf{Section \ref{sec:more_sink}} provides additional visualizations of the attention sink phenomenon. By visualizing attention heatmaps from the vision encoder, we further substantiate the motivation behind our decoupled spatial token selection strategy.
    
    \item \textbf{Section \ref{sec:pseudocode}} details the implementation of our framework, providing the pseudocode for the two core components: the \textit{inner-vision encoder frame merging} and the \textit{decoupled spatial token selection}.

    \item \textbf{Section \ref{sec:future-work}} presents the future work of EarlyTom, including potential directions for system co-design, heterogeneous inference optimization, and acceleration for the decoding stage in multimodal models.
\end{itemize}

\section{Generalizability Analysis on LLaVA-Video and Qwen2.5-VL}
\label{sec:exp}

To further verify the effectiveness and broad applicability of our framework, we extend our evaluation to the LLaVA-Video-7B model and Qwen2.5-VL-7B model.

\paragraph{Efficiency analysis.} As detailed in Table~\ref{tab:vid_benchmark}, EarlyTom consistently delivers substantial improvements in computational efficiency across all tested token retention settings. By performing frame merging directly within the vision encoder, our method effectively reduces the prefilling FLOPs. For instance, at a 15\% retention rate, EarlyTom reduces the FLOPs ratio to 35.1\% and achieves a time-to-first-token of 947.4 ms, representing a 6.8$\times$ speedup compared to the full-token baseline (6429.3 ms).
The efficiency advantages are also corroborated on the Qwen2.5-VL-7B backbone (Table~\ref{tab:qwen25vl}). Specifically, while trivial baselines like Average Pooling and Uniform Subsampling result in a 16.6\% FLOPs ratio, EarlyTom further optimizes this to 12.2\% (67.7T), achieving a significantly faster TTFT (3667 ms) than both the full model and the native token reduction baselines.

\paragraph{Accuracy and trade-off.} Table~\ref{tab:vid_benchmark} presents a comprehensive comparison of accuracy and efficiency. EarlyTom maintains competitive performance on standard video understanding benchmarks, achieving an average score of 56.43\% while operating with significantly reduced computational overhead. These results demonstrate that EarlyTom can successfully generalize to the LLaVA-Video architecture, providing an efficient inference solution that balances high throughput with reliable model performance.
This robust generalizability is further evidenced by our results on the Qwen2.5-VL-7B backbone (Table~\ref{tab:qwen25vl}). At a 15\% token retention ratio, EarlyTom achieves an average score of 62.2\%, which significantly outperforms the Uniform Subsampling and Average Pooling baselines. Notably, EarlyTom maintains higher accuracy than these trivial baselines while utilizing even fewer FLOPs, demonstrating a superior Pareto frontier in the accuracy-efficiency trade-off.

\paragraph{Ablation studies.} To investigate the individual contribution of our proposed modules, we conduct fine-grained ablation studies on the Qwen2.5-VL architecture, as summarized in Table~\ref{tab:qwen25vl}. We observe that both decoupled spatial token selection and weighted frame merging are essential for maintaining optimal performance under aggressive compression. Specifically, removing the spatial selection module leads to a performance drop from 62.2\% to 61.4\%, indicating its critical role in identifying and preserving informative regions across frames. Similarly, excluding the weighted merging strategy results in a decline to 61.3\%, underscoring the importance of our importance-aware aggregation. These results confirm that the synergy between spatial selection and temporal merging is the key to EarlyTom’s ability to preserve high-fidelity visual information while reducing token redundancy.
\begin{table*}[t]
    \centering
    \caption{\textbf{Efficiency comparison with SoTA methods on the LLaVA-Video-7B model.} \textbf{Best} results are in bold, \underline{second-best} results are underlined. Time-to-first-token is denoted as TTFT for simplicity. All efficiency results are measured on a single NVIDIA A100 GPU.}
    \setlength{\tabcolsep}{0.8mm}
    \resizebox{1.0\linewidth}{!}
    {
        \begin{tabular}{c|l|ccccc|cccc|cc}
        \toprule
        \multirow{2}{*}{Model} &
        \multirow{2}{*}{Method} &
        \multirow{2}{*}{\shortstack{Before LLM\\Retained Ratio}} &
        \multirow{2}{*}{\shortstack{Prefilling\\FLOPs (T) $\downarrow$}} &
        \multirow{2}{*}{\shortstack{FLOPs\\Ratio $\downarrow$}} &
        \multirow{2}{*}{\shortstack{TTFT\\(ms) $\downarrow$}} &
        \multirow{2}{*}{\shortstack{Throughput\\(tokens/s) $\uparrow$}} &
        \multirow{2}{*}{\shortstack{MVBench\\$\uparrow$}} &
        \multirow{2}{*}{\shortstack{EgoSchema\\$\uparrow$}} &
        \multirow{2}{*}{\shortstack{LongVideo\\Bench $\uparrow$}} &
        \multirow{2}{*}{\shortstack{VideoMME\\$\uparrow$}} &
        \multicolumn{2}{c}{Avg. $\uparrow$} \\
         & & & & & & & & & & & Score & \% \\
        \midrule
        \multirow{6}{*}{\shortstack{LLaVA- \\ Video-7B}}
        & \cellcolor{gray!20} Vanilla & \cellcolor{gray!20} 100\% & \cellcolor{gray!20} 246.2 & \cellcolor{gray!20} 100\% & \cellcolor{gray!20} 6429.3 & \cellcolor{gray!20} 8.1 & \cellcolor{gray!20} 60.4 & \cellcolor{gray!20} 57.2 & \cellcolor{gray!20} 58.9 & \cellcolor{gray!20} 64.3 & \cellcolor{gray!20} 60.2 & \cellcolor{gray!20} 100 \\
        & \cellcolor{white} FastV~\citep{chen2024fastv}$_{\text{ECCV'24}}$ & \cellcolor{white} 100\% & \cellcolor{white} 158.2 & \cellcolor{white} 64.3\% & \cellcolor{white} 3494.3 & \cellcolor{white} 10.0 & \cellcolor{white} 54.3 & \cellcolor{white} 54.1 & \cellcolor{white} 55.0 & \cellcolor{white} 58.8 & \cellcolor{white} 55.6 & \cellcolor{white} 92.4 \\
        & \cellcolor{white} PyramidDrop~\citep{xing2025pyramiddrop}$_{\text{CVPR'25}}$ & \cellcolor{white} 100\% & \cellcolor{white} 159.4 & \cellcolor{white} 64.7\% & \cellcolor{white} 3494.8 & \cellcolor{white} 10.1 & \cellcolor{white} 55.9 & \cellcolor{white} 54.3 & \cellcolor{white} 54.7 & \cellcolor{white} 61.9 & \cellcolor{white} 56.7 & \cellcolor{white} 94.2 \\
        & \cellcolor{white} VisionZip~\citep{yang2025visionzip}$_{\text{CVPR'25}}$ & \cellcolor{white} 15\% & \cellcolor{white} 159.4 & \cellcolor{white} 64.7\% & \cellcolor{white} 3241.4 & \cellcolor{white} 14.2 & \cellcolor{white} 56.7 & \cellcolor{white} 54.7 & \cellcolor{white} 54.7 & \cellcolor{white} 60.7 & \cellcolor{white} 56.7 & \cellcolor{white} 94.2 \\
        & \cellcolor{white} HoliTom~\citep{shao2025holitom}$_{\text{NeurIPS'25}}$ & \cellcolor{white} 15\% & \cellcolor{white} \underline{156.6} & \cellcolor{white} \underline{63.6\%} & \cellcolor{white} \underline{1669.5} & \cellcolor{white} \textbf{17.1} & \cellcolor{white} 57.7 & \cellcolor{white} 54.8 & \cellcolor{white} 56.2 & \cellcolor{white} 62.1 & \cellcolor{white} 57.7 & \cellcolor{white} 95.8 \\
        & \cellcolor{lightblue} \textbf{EarlyTom} & \cellcolor{lightblue} 15\% & \cellcolor{lightblue} \textbf{86.4} & \cellcolor{lightblue} \textbf{35.1\%} & \cellcolor{lightblue} \textbf{947.4} & \cellcolor{lightblue} \underline{16.7} & \cellcolor{lightblue} 55.8 & \cellcolor{lightblue} 54.7 & \cellcolor{lightblue} 53.9 & \cellcolor{lightblue} 61.3 & \cellcolor{lightblue} 56.4 & \cellcolor{lightblue} 93.7 \\
        \bottomrule
        \end{tabular}
    }
\label{tab:vid_benchmark}
\vspace{-2.mm}
\end{table*}

\begin{table*}[t]
    \caption{\textbf{Experiment results on trivial baselines and ablation studies.} All results are obtained on Qwen2.5-VL-7B with a maximum of 768 frames and a retain ratio of 15\%. Efficiency metrics are measured under a 23k-token context length on a single NVIDIA A100 GPU.
    }
    \setlength{\tabcolsep}{4.2mm}
    \resizebox{1.0\linewidth}{!}{
        \begin{tabular}{l|ccc|ccccc|c}
        \toprule
        \multirow{2}{*}{Method}
        & \multirow{2}{*}{\shortstack{Prefilling\\FLOPs (T) $\downarrow$}} 
        & \multirow{2}{*}{\shortstack{FLOPs\\Ratio $\downarrow$}} 
        & \multirow{2}{*}{\shortstack{TTFT\\(ms) $\downarrow$}} 
        & \multirow{2}{*}{\shortstack{MVBench $\uparrow$}} 
        & \multicolumn{4}{c|}{VideoMME $\uparrow$} 
        & \multirow{2}{*}{\shortstack{Avg. \\ Score}} \\
        \cline{6-9}
        & & & & & Short & Medium & Long & Average \\
        \midrule

        \rowcolor{gray!20}
        Qwen2.5-VL-7B
            & 554.7
            & 100\%
            & 6842
            & 67.1
            & 76.0
            & 66.0
            & 55.1
            & 65.7
            & 66.4 \\
        \midrule

        Average Pooling
            & 91.9
            & 16.6\%
            & 4609
            & 56.8
            & 66.4
            & 57.3 
            & 51.1 
            & 58.3
            & 57.6 \\

        Uniform Subsampling
            & \underline{91.9}
            & \underline{16.6\%}
            & \underline{4578}
            & 57.7
            & 68.6
            & 59.6 
            & \textbf{55.0}
            & 60.8
            & 59.3 \\

        EarlyTom w/o Decoupled Spatial Token Selection
            & 67.7
            & 12.2\%
            & 3667
            & \underline{60.7}
            & \textbf{71.0}
            & \underline{61.6}
            & \underline{53.6}
            & \textbf{62.0}
            & \underline{61.4} \\

        EarlyTom w/o Weighted Frame Merging
            & 67.7
            & 12.2\%
            & 3667
            & \underline{60.7}
            & 70.5
            & \textbf{62.3} 
            & 52.7
            & 61.8
            & 61.3 \\

        \rowcolor{lightblue}
        \textbf{EarlyTom} 
            & \textbf{67.7}
            & \textbf{12.2\%}
            & \textbf{3667}
            & \textbf{62.5}
            & \underline{70.7}
            & \underline{61.6}
            & \underline{53.6} 
            & \underline{61.9}
            & \textbf{62.2} \\

        \bottomrule
        \end{tabular}
    }
    \label{tab:qwen25vl}
\vspace{-2.mm}
\end{table*} 

\begin{table}[t]
    \centering
    \caption{Details of the hyperparameters on LLaVA-OneVision.}
    \setlength{\tabcolsep}{1.5mm}
    \resizebox{1.0\linewidth}{!}{
    \begin{tabular}{c|cccccc}
        \toprule
        \multirow{2}{*}{\shortstack{Retained \\ Ratio}} &
        \multirow{2}{*}{\shortstack{w. Inner\\LLM}} &
        \multirow{2}{*}{\shortstack{EMA factor\\ $\alpha$}} &
        \multirow{2}{*}{\shortstack{MVBench\\$\uparrow$}} &
        \multirow{2}{*}{\shortstack{EgoSchema\\$\uparrow$}} &
        \multirow{2}{*}{\shortstack{LongVideo\\Bench $\uparrow$}} &
        \multirow{2}{*}{\shortstack{VideoMME\\$\uparrow$}} \\
        & & & & & \\
        \midrule
        \multicolumn{7}{c}{\textbf{LLaVA-OneVision-7B}} \\
        \midrule
        \multirow{2}{*}{25\%} & \multirow{2}{*}{$\cmark$} & \multirow{2}{*}{0.9} & $\tau_\text{seg}$=0.8  & $\tau_\text{seg}$=0.7 & $\tau_\text{seg}$=0.8 & $\tau_\text{seg}$=0.6 \\
        & & & [6,14,20] & [10,21,23] & [6,21,23] & [10,21,23] \\
        \midrule
        \multirow{2}{*}{20\%} & \multirow{2}{*}{$\cmark$} & \multirow{2}{*}{0.9} & $\tau_\text{seg}$=0.8 & $\tau_\text{seg}$=0.6 & $\tau_\text{seg}$=0.6 & $\tau_\text{seg}$=0.5 \\
        & & & [6,14,20] & [10,21,23] & [10,21,23] & [8,21,23] \\
        \midrule
        \multirow{2}{*}{15\%} & \multirow{2}{*}{$\cmark$} & \multirow{2}{*}{0.9} & $\tau_\text{seg}$=0.8 & $\tau_\text{seg}$=0.5 & $\tau_\text{seg}$=0.5 & $\tau_\text{seg}$=0.4 \\
        & & & [6,14,20] & [10,21,23] & [10,21,23] & [8,21,23] \\
        \midrule
        \multirow{2}{*}{10\%} & \multirow{2}{*}{$\cmark$} & \multirow{2}{*}{0.9} & $\tau_\text{seg}$=0.65 & $\tau_\text{seg}$=0.3 & $\tau_\text{seg}$=0.3 & $\tau_\text{seg}$=0.3 \\
        & & & [8,14,20] & [10,21,23] & [10,21,23] & [10,21,23] \\
        \midrule
        \multicolumn{7}{c}{\textbf{LLaVA-OneVision-0.5B}} \\
        \midrule
        \multirow{2}{*}{25\%} & \multirow{2}{*}{$\cmark$} & \multirow{2}{*}{0.9} & $\tau_\text{seg}$=0.8 & $\tau_\text{seg}$=0.7 & $\tau_\text{seg}$=0.8 & $\tau_\text{seg}$=0.6 \\
        & & & [8,21,23] & [10,21,23] & [6,21,23] & [8,21,23] \\
        \midrule
        \multirow{2}{*}{20\%} & \multirow{2}{*}{$\cmark$} & \multirow{2}{*}{0.9} & $\tau_\text{seg}$=0.8 & $\tau_\text{seg}$=0.6 & $\tau_\text{seg}$=0.6 & $\tau_\text{seg}$=0.5 \\
        & & & [8,21,23] & [10,21,23] & [10,21,23] & [8,21,23] \\
        \midrule
        \multirow{2}{*}{15\%} & \multirow{2}{*}{$\cmark$} & \multirow{2}{*}{0.9} & $\tau_\text{seg}$=0.8 & $\tau_\text{seg}$=0.5 & $\tau_\text{seg}$=0.5 & $\tau_\text{seg}$=0.4 \\
        & & & [8,21,23] & [10,21,23] & [10,21,23] & [8,21,23] \\
        \midrule
        \multirow{2}{*}{10\%} & \multirow{2}{*}{$\cmark$} & \multirow{2}{*}{0.9} & $\tau_\text{seg}$=0.65 & $\tau_\text{seg}$=0.3 & $\tau_\text{seg}$=0.3 & $\tau_\text{seg}$=0.3 \\
        & & & [8,21,23] & [10,21,23] & [10,21,23] & [8,21,23] \\
        \bottomrule
    \end{tabular}
    }
    \label{tab:hyper}
    \vspace{-6.mm}
\end{table}

\section{Detailed Analysis of TTFT Latency Decomposition}
\label{sec:more_ttft}

\begin{figure*}
    \centering
    \begin{subfigure}[b]{0.45\linewidth}
        \centering
        \includegraphics[width=\textwidth]{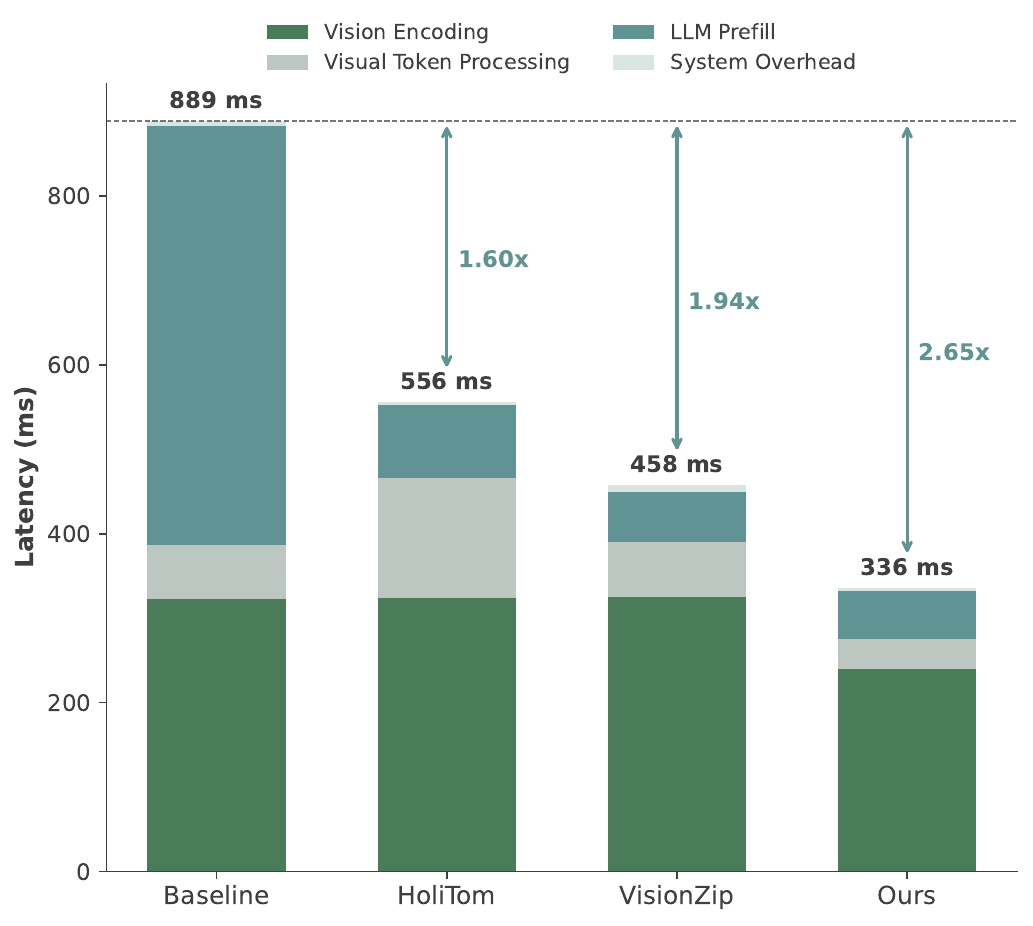}
        \subcaption{10\% token retention rate.}
        \label{fig:sub1}
    \end{subfigure}
    \hfill 
    \begin{subfigure}[b]{0.45\linewidth}
        \centering
        \includegraphics[width=\textwidth]{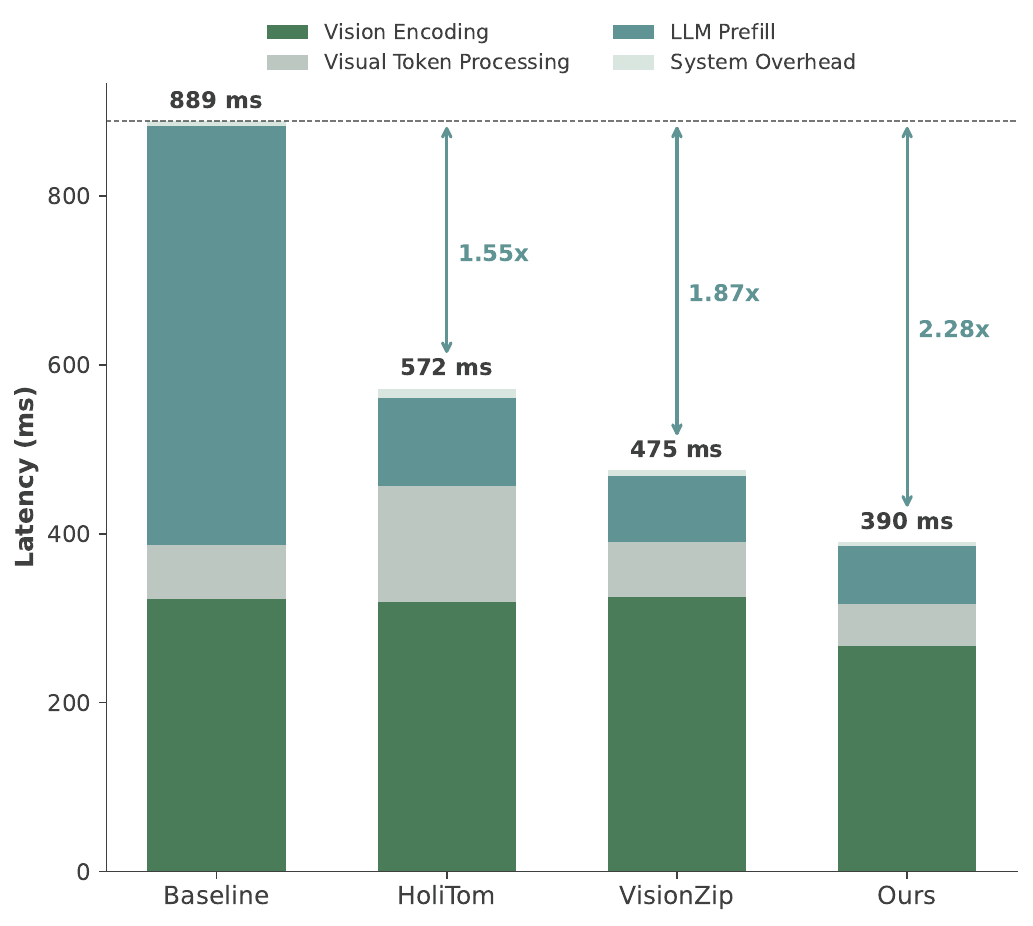}
        \subcaption{15\% token retention rate.}
        \label{fig:sub2}
    \end{subfigure}
    
    \begin{subfigure}[b]{0.45\linewidth}
        \centering
        \includegraphics[width=\textwidth]{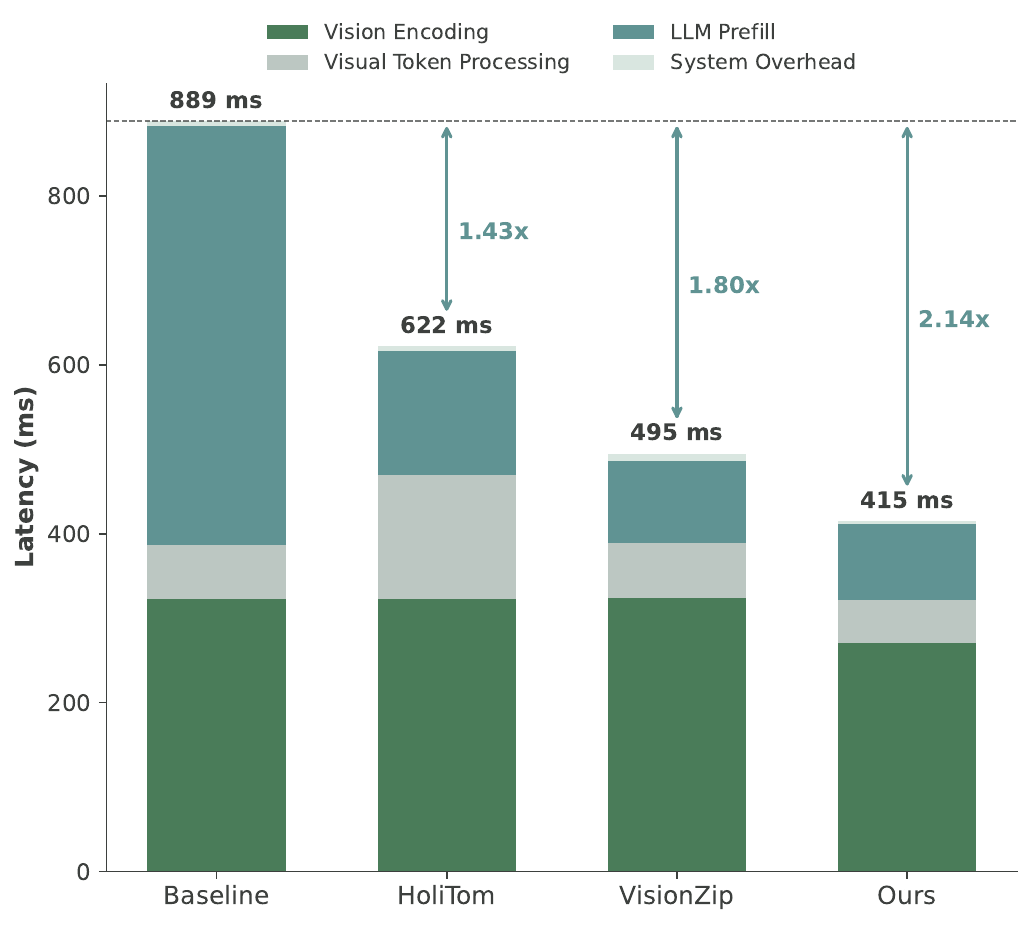}
        \subcaption{20\% token retention rate.}
        \label{fig:sub3}
    \end{subfigure}
    \hfill
    \begin{subfigure}[b]{0.45\linewidth}
        \centering
        \includegraphics[width=\textwidth]{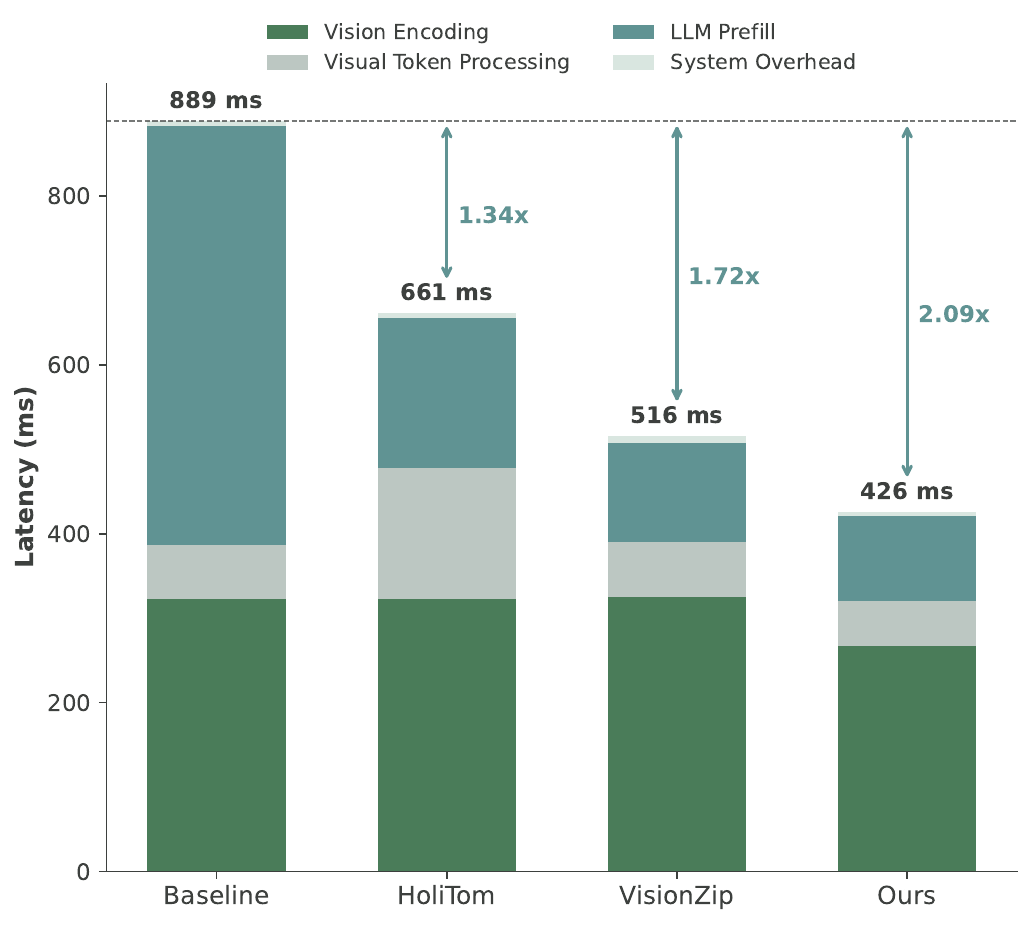}
        \subcaption{25\% token retention rate.}
        \label{fig:sub4}
    \end{subfigure}
    
    \caption{Time-to-first-token (TTFT) comparison on the \textbf{LLaVA-OneVision-7B} model. We report the latency breakdown (vision encoding, token processing, LLM prefill, and system overhead) across different methods.}
    \label{fig:7b_results}
\end{figure*}

\begin{figure*}
    \centering
    \begin{subfigure}[b]{0.45\linewidth}
        \centering
        \includegraphics[width=\textwidth]{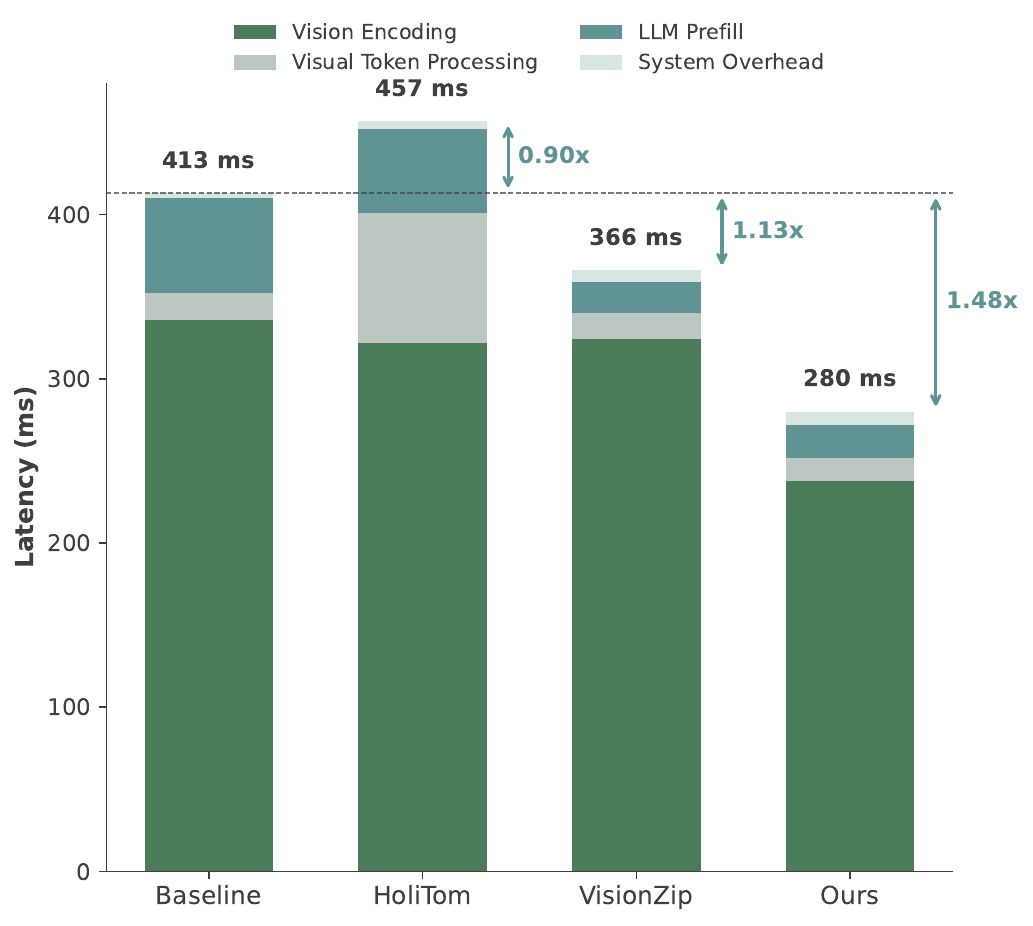}
        \subcaption{10\% token retention rate.}
        \label{fig:sub5}
    \end{subfigure}
    \hfill
    \begin{subfigure}[b]{0.45\linewidth}
        \centering
        \includegraphics[width=\textwidth]{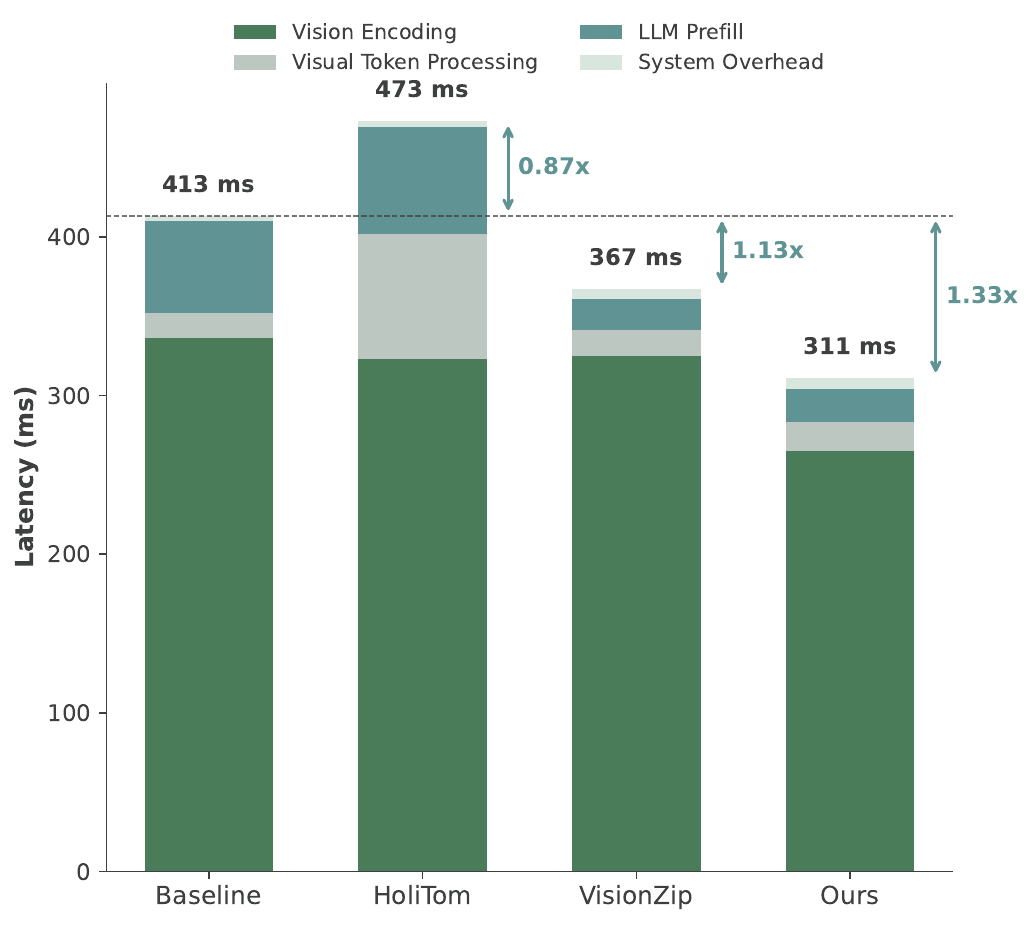}
        \subcaption{15\% token retention rate.}
        \label{fig:sub6}
    \end{subfigure}
    
    \begin{subfigure}[b]{0.45\linewidth}
        \centering
        \includegraphics[width=\textwidth]{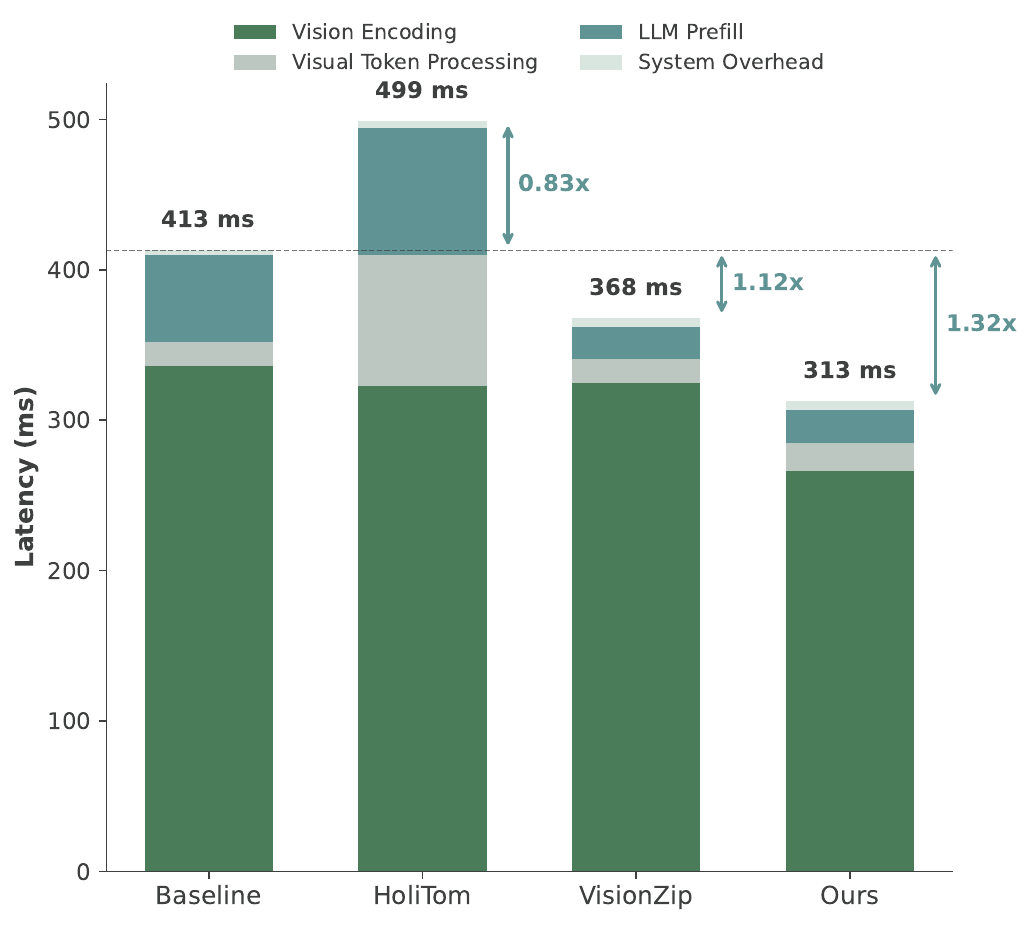}
        \subcaption{20\% token retention rate.}
        \label{fig:sub7}
    \end{subfigure}
    \hfill
    \begin{subfigure}[b]{0.45\linewidth}
        \centering
        \includegraphics[width=\textwidth]{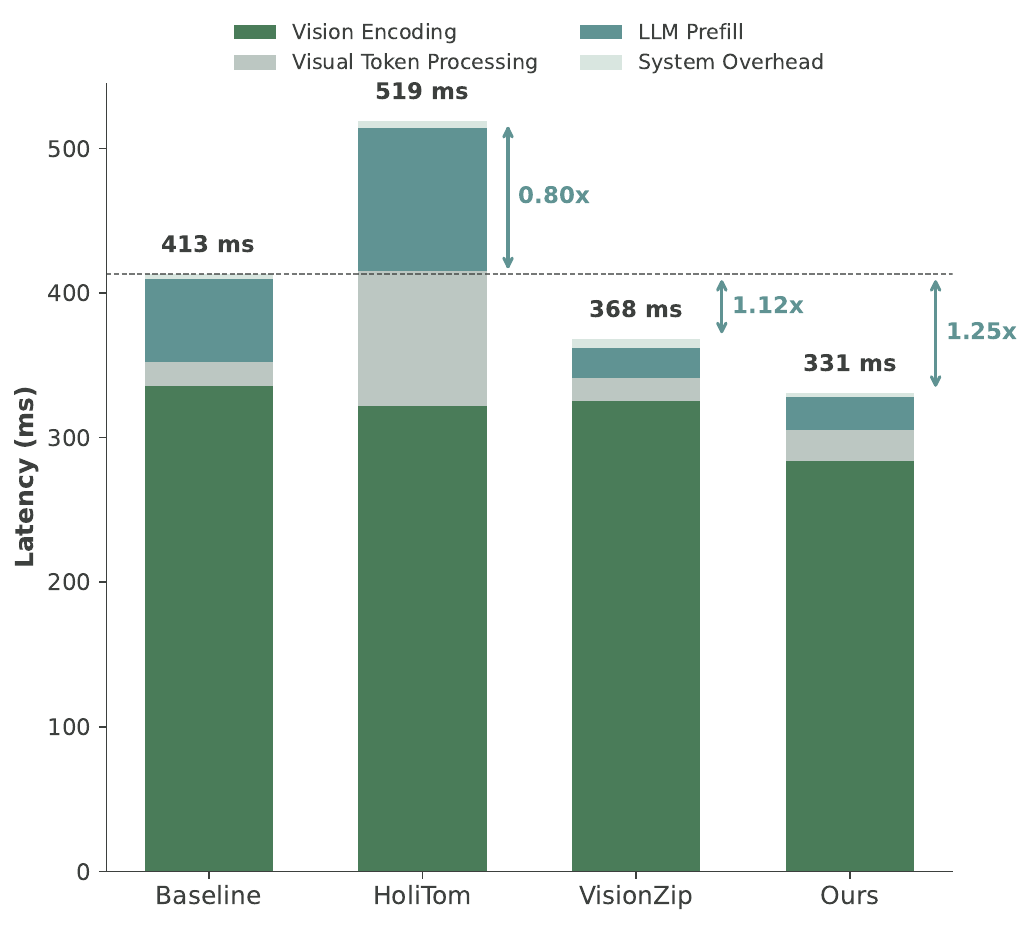}
        \subcaption{25\% token retention rate.}
        \label{fig:sub8}
    \end{subfigure}
    
    \caption{Time-to-first-token (TTFT) comparison on the \textbf{LLaVA-OneVision-0.5B} model. We report the latency breakdown (vision encoding, token processing, LLM prefill, and system overhead) across different methods.}
    \label{fig:0.5b_results}
\end{figure*}

In this section, we provide a fine-grained visualization of the Time-to-First-Token (TTFT) latency composition for both LLaVA-OneVision-7B (Figure~\ref{fig:7b_results}) and LLaVA-OneVision-0.5B (Figure~\ref{fig:0.5b_results}) under varying token retention rates (10\%, 15\%, 20\%, and 25\%). The total latency is decomposed into four components: Vision Encoding, Visual Token Processing, LLM Prefill, and System Overhead.

\paragraph{Analysis on LLaVA-OneVision-7B.} As illustrated in Figure~\ref{fig:7b_results}, the vision encoding stage constitutes a dominant portion of the total latency for the Baseline, HoliTom, and VisionZip. While existing methods like HoliTom and VisionZip effectively reduce the LLM prefill latency through token reduction, they fail to address the high computational cost of the vision encoder. Moreover, HoliTom introduces significant computational overhead during the Visual Token Processing stage, which partially offsets the gains from reduced prefill time. In contrast, EarlyTom directly compresses redundancy within the vision encoder, achieving a substantial reduction in encoding latency. Consequently, our method achieves the lowest total TTFT across all settings, delivering a speedup of up to 2.65$\times$ compared to the baseline at a 10\% retention rate.

\paragraph{Analysis on LLaVA-OneVision-0.5B.} The advantages of our approach are consistent across model scales. Figure~\ref{fig:0.5b_results} presents the results on the smaller 0.5B backbone. A notable observation is that on this lightweight model, the computational overhead introduced by comparison methods becomes more detrimental. Specifically, HoliTom exhibits a higher total latency than the Baseline (e.g., 0.90$\times$ speedup at 10\% retention) because the time saved in the LLM prefill stage is insufficient to outweigh the extra cost of its token processing module. Conversely, EarlyTom maintains its superiority by minimizing both vision encoding time and processing overhead. Even with the smaller potential for prefill acceleration in the 0.5B model, our method achieves a robust speedup of 1.48$\times$ (at 10\% retention), validating the effectiveness of our early-stage compression strategy.

\section{Visualization of the Attention Sink Phenomenon Across Diverse Video Samples}
\label{sec:more_sink}

\begin{figure*}[ht]
    \centering
    \includegraphics[width=1.0\linewidth]{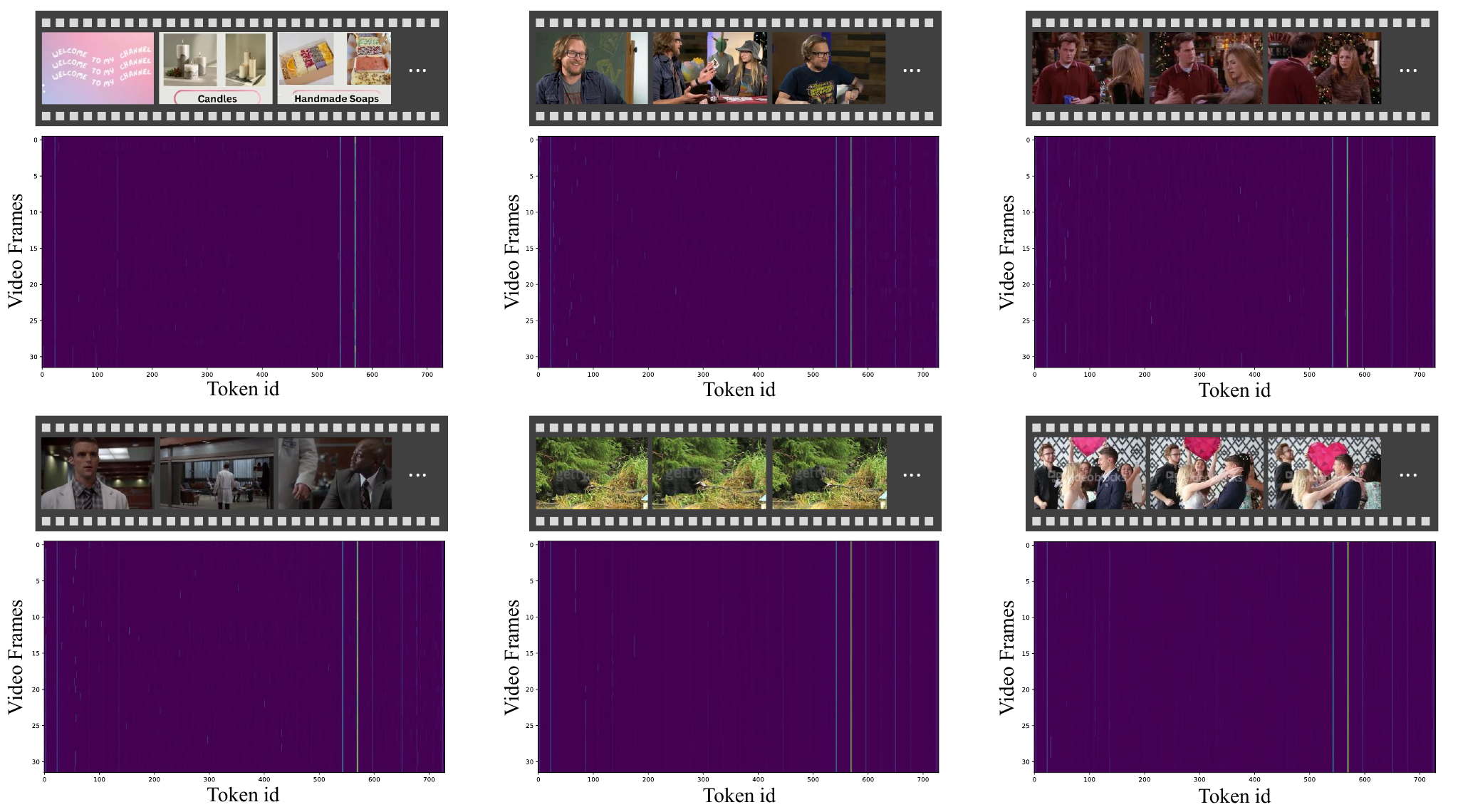}
    \caption{Additional visualizations of attention score distributions. We present the attention heatmaps from the SigLIP vision encoder across six randomly selected videos. The consistent vertical stripes (highlighted in bright colors) indicate that specific spatial tokens accumulate disproportionately high attention scores throughout the temporal sequence. This observation confirms that attention ``sinks" are a widely existing structural characteristic in the vision encoder, motivating the design of our decoupled token selection strategy.}
    \label{fig:sup_attn_sink}
\end{figure*}

In this section, we provide additional visualizations to further substantiate the analysis of the ``Attention Sink" phenomenon discussed in the main paper. Figure~\ref{fig:sup_attn_sink} displays the attention heatmaps extracted from the SigLIP vision encoder across a diverse set of video samples.

\paragraph{Observation.} A consistent pattern emerges across all examples: distinct vertical stripes appear in the heatmaps, indicating that certain spatial tokens maintain exceptionally high attention scores throughout the entire video sequence. These tokens, often referred to as ``attention sinks," act as static attractors within the feature space, dominating the attention distribution regardless of the changing visual content in dynamic frames.

\paragraph{Motivation for Our Design.} This visualization highlights a critical insight for token compression: simply ranking tokens by attention magnitude might bias the selection towards these static sink tokens, potentially overlooking less prominent but semantically rich dynamic features. Recognizing this inherent distribution characteristic, EarlyTom adopts a Decoupled Spatial Token Selection strategy. By distinguishing between static frames (where sinks are stable) and dynamic frames, and applying tailored selection mechanisms for each, our method ensures that the compressed token set preserves both the necessary structural information (sinks) and the crucial motion-sensitive details, leading to a more robust and balanced video representation.

% Algorithm 1
\begin{algorithm}[ht]
\caption{Inner-Vision Encoder Frame Merging}
\label{alg:inner_encoder_frame_merge}
\begin{algorithmic}

\State \textbf{Input:} Frame features $F \in \mathbb{R}^{B \times L \times D}$, hyperparameters $\alpha, \tau_\text{seg}, \tau_\text{merge}$.
\State \textbf{Output:} Merged frame features $\hat{F}_\text{out} \in \mathbb{R}^{N \times L \times D}$.

\State {\color{forestgreen}{\textit{Streaming Frame Segmentation in~\Cref{eq:ema_seg}}}}
\State $\mathcal{S} \leftarrow \text{SegmentBySimilarity}(F, \alpha, \tau_\text{seg})$, \ $F_\text{merged\_list} \leftarrow [\,]$

\State \textbf{for} each segment $S_\text{seg} = \{F_0, \dots, F_k\}$ in $\mathcal{S}$ \textbf{do}
\State \quad $F_\text{mid} \leftarrow [\,]$, $i \leftarrow 1$

\State \quad {\color{forestgreen}{\textit{Iterate over Middle Frames within the Segment}}}
\State \quad \textbf{while} $i < k$ \textbf{do}
\State \quad \quad {\color{forestgreen}{\textit{Compute Pairwise Frame Similarities}}}
\State \quad \quad $s_i \leftarrow \text{Sim}(F_i, F_{i+1})$, $s_{i+1} \leftarrow \text{Sim}(F_{i+1}, F_{i+2})$

\State \quad \quad {\color{forestgreen}{\textit{Middle Frame Merge Condition in~\Cref{eq:merge_condition}}}}
\State \quad \quad \textbf{if} $s_i > \tau_\text{merge}$ \textbf{and} $s_i > s_{i+1}$ \textbf{then}
\State \quad \quad \quad {\color{forestgreen}{\textit{Weighted Frame Merge in~\Cref{eq:weighted_merge}}}}
\State \quad \quad \quad $\hat{F}_{m} \leftarrow \text{WeightedMerge}(s_{i}, F_{i}, s_{i+1}, F_{i+1})$
\State \quad \quad \quad $F_\text{mid}.\text{append}(\hat{F}_m)$; $i \leftarrow i + 2$

\State \quad \quad \textbf{else}
\State \quad \quad \quad $F_\text{mid}.\text{append}(F_i)$; $i \leftarrow i + 1$
\State \quad \quad \textbf{end if}

\State \quad \textbf{end while}

\State \quad {\color{forestgreen}{\textit{Assemble Merged Segment}}}
\State \quad $F_\text{seg\_out} \leftarrow \text{Concat}(F_0, F_\text{mid}, F_k)$
\State \quad $F_\text{merged\_list}.\text{append}(F_\text{seg\_out})$
\State \textbf{end for}

\State {\color{forestgreen}{\textit{Concatenate All Merged Segments}}}
\State $\hat{F}_\text{out} \leftarrow \text{Concatenate}(F_\text{merged\_list})$
\State \textbf{Return} $\hat{F}_\text{out}$

\end{algorithmic}
\end{algorithm}

\section{Pseudocode of EarlyTom}
\label{sec:pseudocode}

In this section, we provide the detailed pseudocode for the two core components of EarlyTom to facilitate implementation. Algorithm~\ref{alg:inner_encoder_frame_merge} outlines the inner-vision encoder frame merging process, which performs adaptive streaming segmentation and weighted merging to reduce temporal redundancy. Algorithm~\ref{alg:decoupled_spatial_selection} illustrates the decoupled spatial token selection strategy, describing how dynamic and static frames are processed via distinct selection mechanisms to ensure balanced spatial information preservation.

\section{Future Work}
\label{sec:future-work}
EarlyTom reveals that the inference budget is mainly dominated by the prefill stage in VLMs. Although existing methods~\citep{liu2025video, liu2025mixing, liu2025global, chen2025variation, xiong2025prune2drive, zhu2025obs, du2025whichheads, kong2026omni} have proposed various techniques for efficient inference, they primarily focus on algorithm-level improvements rather than system-level optimizations. How to jointly leverage system design and algorithmic techniques in a heterogeneous manner remains an open problem. Meanwhile, recent reasoning models~\citep{feng2025efficient} have exhibited strong scene understanding capabilities, yet they still suffer from lengthy generation steps during the decoding stage. Therefore, accelerating inference and improving efficiency via system–algorithm co-design is essential and worthy of further exploration.
% Algorithm 2
\begin{algorithm}[t]
\caption{Decoupled Spatial Token Selection}
\label{alg:decoupled_spatial_selection}
\begin{algorithmic}

\State \textbf{Input:} Features $\hat{F}$ and attentions $A$ from vision encoder, segment list $\mathcal{S}$, target ratio $r$.
\State \textbf{Output:} Final compressed features $\hat{\hat{F}}$.

\State {\color{forestgreen}{\textit{Decouple Frames into Dynamic and Static Sets}}}
\State $\hat{F}^d, A^d \leftarrow [\,]$, $[\,]$
\State $\hat{F}^s, A^s \leftarrow [\,]$, $[\,]$
\State \textbf{for} each segment $S_\text{seg} = \{\hat{F}_0, \dots, \hat{F}_k\}$ in $\mathcal{S}$ \textbf{do}
\State \quad $\hat{F}^d.\text{append}(\hat{F}_0, \hat{F}_k)$; \ $A^d.\text{append}(A_0, A_k)$
\State \quad $\hat{F}^s.\text{append}(\hat{F}_{1:k-1})$; \ $A^s.\text{append}(A_{1:k-1})$
\State \textbf{end for}

\State {\color{forestgreen}{\textit{Compute Re-scaled Retention Ratio in~\Cref{eq:r_hat}}}}
\State $\hat{r}  = \frac{r}{(\frac{B-N}{B})*L}$

\State {\color{forestgreen}{\textit{Compress Dynamic Frames via Global Top-K Selection}}}
\State $\hat{\hat{F}}^d \leftarrow \text{GlobalTopKSelection}(\hat{F}^d, A^d, \hat{r})$

\State {\color{forestgreen}{\textit{Compress Static Frames via Local-window Selection}}}
\State $\hat{\hat{F}}^s \leftarrow \text{LocalWindowSelection}(\hat{{F}}^s, A^s, \hat{r})$

\State {\color{forestgreen}{\textit{Gather and Reorder Selected Tokens in Temporal Order}}}
\State $\hat{\hat{F}} \leftarrow \text{GatherAndReorder}(\hat{\hat{F}}^d, \hat{\hat{F}}^s)$

\State \textbf{Return} $\hat{\hat{F}}$

\end{algorithmic}
\end{algorithm}


\end{document}